\documentclass[journal]{IEEEtran}

\usepackage{cite}
\usepackage{amsmath,amssymb,amsfonts}
\usepackage{algorithmic}
\usepackage{graphicx}
\usepackage{textcomp}
\usepackage{csquotes}
\usepackage{xcolor,soul}
\usepackage{adjustbox}
\usepackage{hyperref}
\usepackage{multirow}
\usepackage{titlesec}
\usepackage{subcaption}
\usepackage{makecell}
\usepackage{algorithmic}
\usepackage[ruled,vlined,linesnumbered,boxed]{algorithm2e}
\usepackage{orcidlink}

\ifCLASSINFOpdf
\else
\fi

\begin{document}

\title{Pedestrian Inertial Navigation: An Overview of Model and Data-Driven Approaches}

\author{Itzik Klein \orcidlink{0000-0001-7846-0654}
\thanks{I. Klein heads the Autonomous Navigation and Sensor Fusion Lab, Hatter Department of Marine Technologies, Charney School of Marine Science, University of Haifa, Haifa 3498838,
Israel (e-mail: kitzik@univ@haifa.ac.il).}
}


\maketitle
\begin{abstract}
The task of indoor positioning is fundamental to several applications, including navigation, healthcare, location-based services, and security. An emerging field is inertial navigation for pedestrians, which relies only on inertial sensors for positioning. In this paper, we present inertial pedestrian navigation models and learning approaches. Among these, are methods and algorithms for shoe-mounted inertial sensors and pedestrian dead reckoning (PDR) with unconstrained inertial sensors. We also address three categories of data-driven PDR strategies: activity-assisted, hybrid approaches, and learning-based frameworks.
\end{abstract}


%
\section{Introduction}
\begin{figure}[htbp]
	\centering
    \includegraphics[width=0.35\textwidth]
    {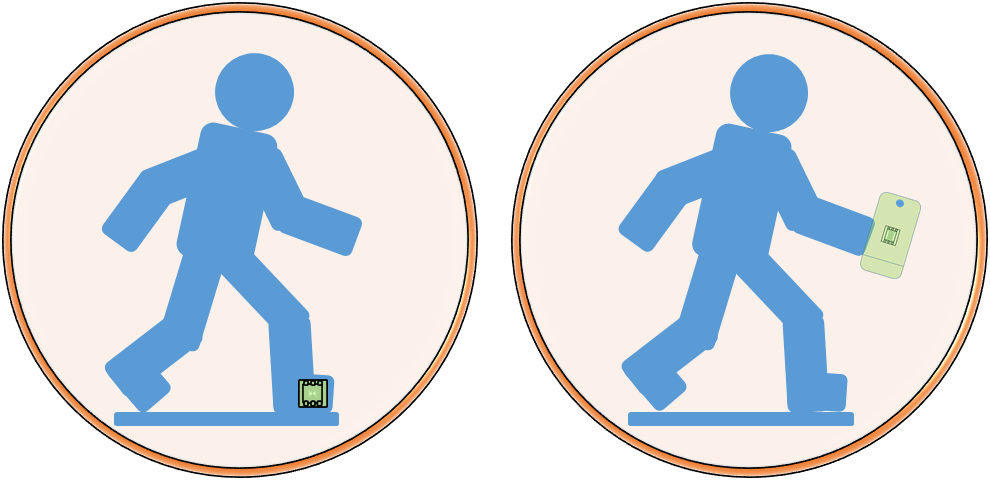}   
    \caption{Shoe-mounted INS (left) requires solving the INS equations and a nonlinear filter with information aiding. The PDR framework (right) is applicable to unconstrained inertial sensors, such as those in smartphones.}\label{Fig:show_and_pdr}
\end{figure}

\noindent
Indoor positioning is a fundamental task for various applications, including navigation, health-monitoring, location-based services, and security. Because global navigation satellite system (GNSS) signals are not available indoors,  positioning relies on other approaches such as WiFi ~\cite{liu2020surveywifi,yang2015wifi,kumar2024h2lwrf} or other radio frequency signals~\cite{brena2017evolution}, visual positioning~\cite{khan2022recent}, floor plan matching~\cite{zhou2023fusion}, and inertial sensing  solutions~\cite{titterton2004strapdown,farrell2008aided}.
Of the latter, several approaches exists, including inertial navigation system (INS), shoe-mounted INS (SM-INS), and  pedestrian dead reckoning (PDR).\\
\noindent
For indoor navigation, low-performance inertial sensors are  commonly used, for example, those installed in smartphones and wearable devices. Therefore, without external updates, the classical INS algorithm, which requires three integrations on the measured inertial data, results in large positioning errors.  To overcome this limitation, inertial sensors can be mounted on a shoe. To limit the inertial drift, zero velocity updates and other methods of information aiding are used. But for unconstrained inertial sensors, like in smartphones, information aiding is not applicable, therefore approaches like PDR are used, which require less integrations on the inertial readings. \\
\noindent
The present paper is focused on SM-INS and PDR approaches, as shown in Figure~\ref{Fig:show_and_pdr}. Section~\ref{sec:modelbasedpdr} presents the model-based PDR framework and explains each part and its role in a common PDR algorithm. Section~\ref{sec:ShoeMountedIMU} describes the SM-INS approach, with all relevant algorithms, including INS, nonlinear filtering, zero velocity detector, and information aiding.  Section~\ref{sec:ddpdr} lists three strategies for data-driven PDR, including activity-assisted, hybrid, and learning-based. Finally, Section~\ref{sec:conc}  summarizes the paper.
\section{Model-Based Pedestrian Dead Reckoning}\label{sec:modelbasedpdr}
\begin{figure}[htbp]
	\centering
    \includegraphics[width=0.4\textwidth]
    {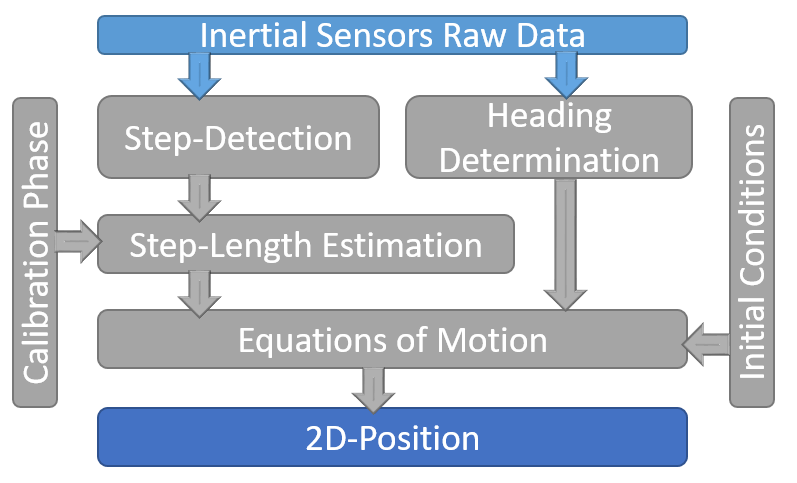}
    \caption{Model-based PDR stages.}\label{Fig:pdr_model}
\end{figure}

\noindent
Inertial sensors, such as accelerometers and gyroscopes, are required for using model-based PDR algorithms. Smartphones or wearable devices with inertial sensors are commonly used for this purpose. An exception is shoe-mounted inertial sensors (discussed below in Section~\ref{sec:ShoeMountedIMU}).\\
Model-based PDR consists of four stages (Figure~\ref{Fig:pdr_model}):
\begin{enumerate}
    \item \textbf{Step detection}: The pedestrian steps are detected generally based on accelerometer readings.
    \item \textbf{Step-length estimation}: Several approaches may be used to estimate pedestrian step-length, including regression-based, biomechanical models, and empirical relationships.
    \item \textbf{Heading determination}: The heading of the pedestrian is estimated from the gyroscope and/or magnetometer readings.
    \item \textbf{Position update}: The current pedestrian position is determined based on initial conditions,  heading angle (stage 3), and step-length size (stage 2).
\end{enumerate}
As with any dead-reckoning method, the user's initial position must be known before undertaking the PDR cycle. Additionally, most step-length algorithms require a calibration phase to determine their parameters. Finally, we note that most model-based PDR approaches assume that the user is walking on the same plane (floor), allowing only the estimation of the pedestrian's 2D position, excluding  altitude. \\
\noindent
In the following subsections, we  elaborate on each stage in the model-based PDR approach   and provide frequently used implementation methods. 
\subsection{Step detection approaches}
\begin{figure}[htbp]
	\centering
    \includegraphics[width=0.5\textwidth]
    {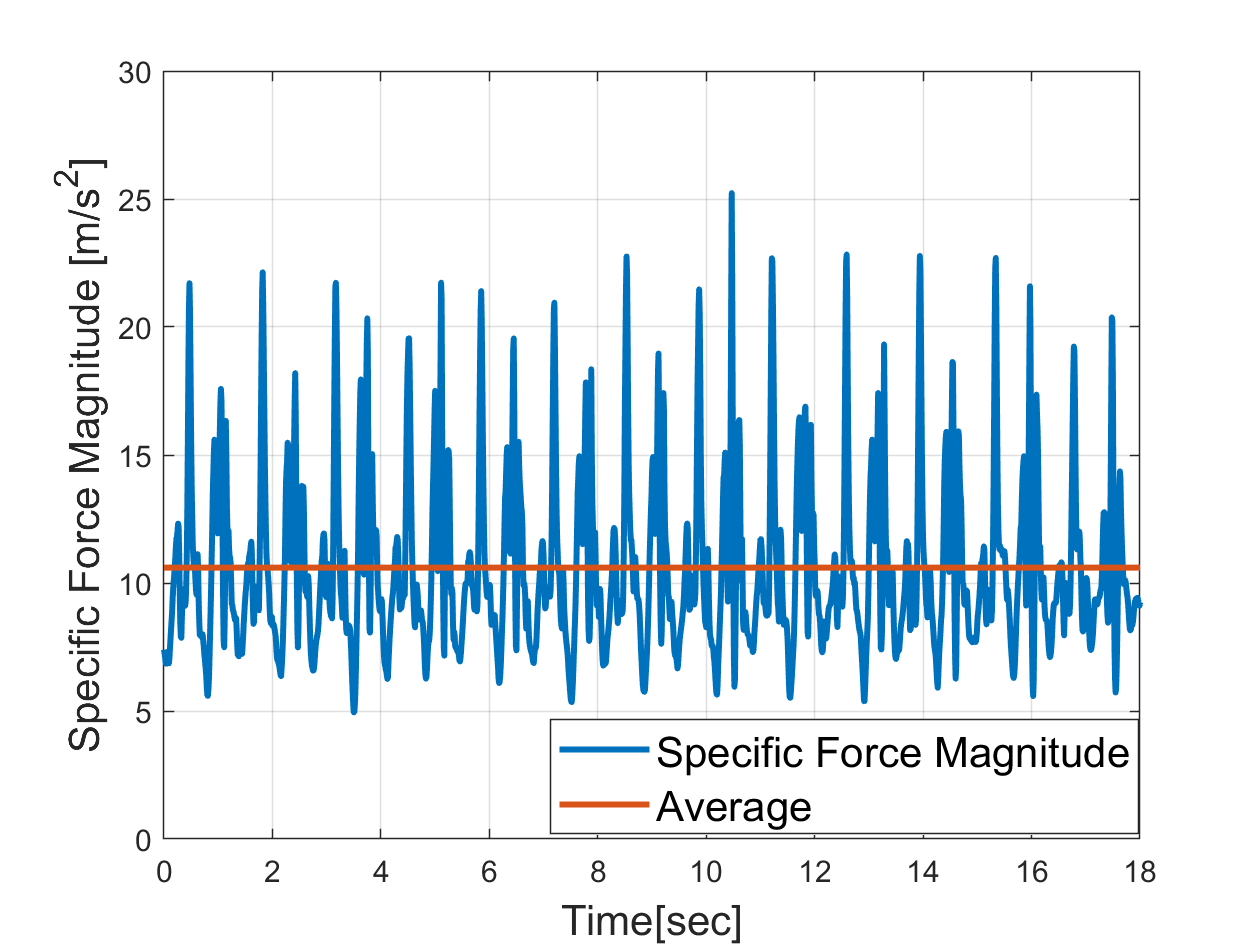}   
    \caption{Specific force magnitude as a function of time and its average value. This recording was taken with a smartphone in texting mode by a pedestrian walking for 25.4 meters.}\label{Fig:accmag}
\end{figure}

\begin{figure}[htbp]
	\centering
    \includegraphics[width=0.5\textwidth]
    {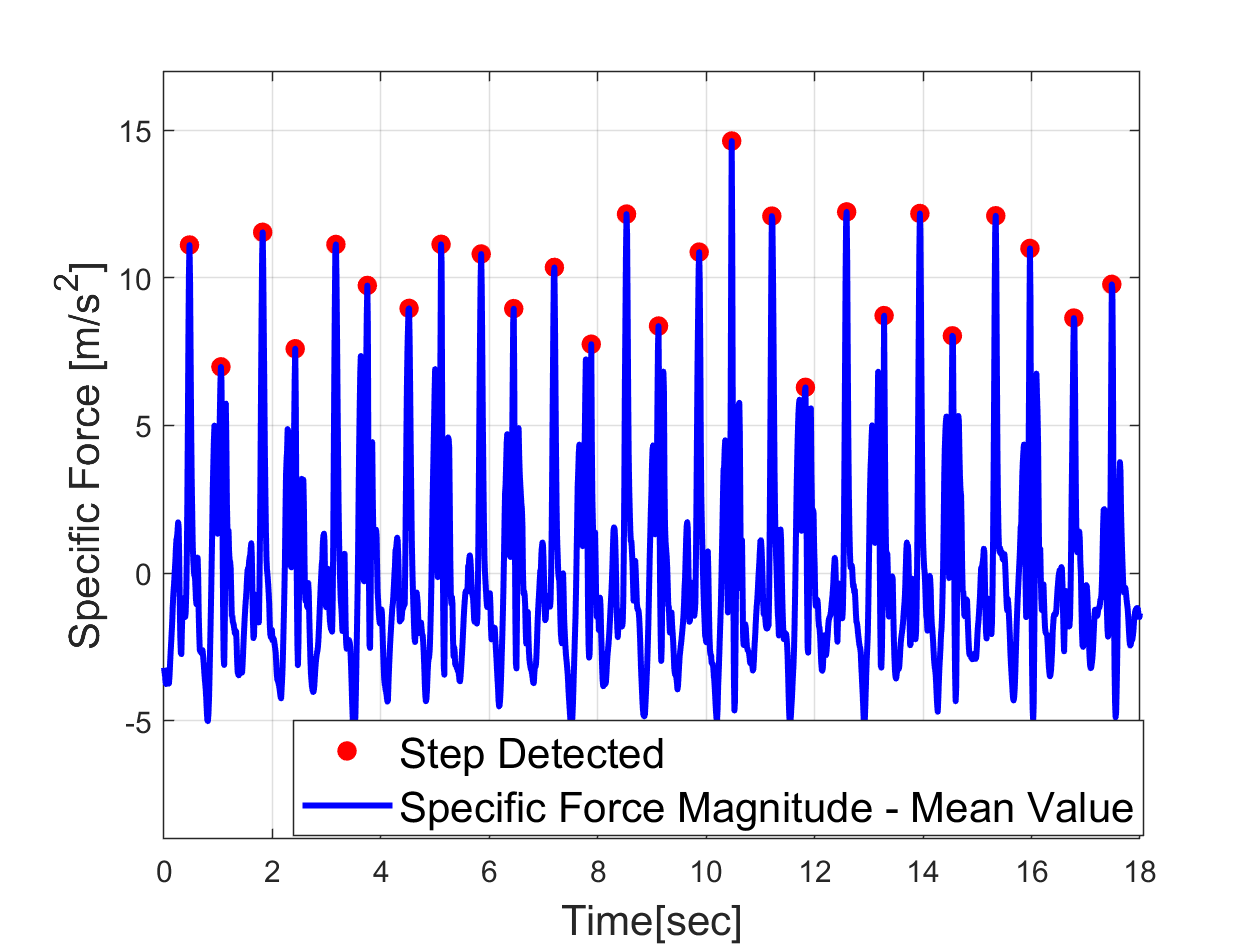}   
    \caption{Identified user steps. This recording was taken with a smartphone in texting mode by a pedestrian walking for 25.4 meters.}\label{Fig:steps}
\end{figure}

\noindent
Steps can be detected in several ways. Although some approaches have been derived for specific inertial sensor locations, with appropriate parameter tuning they can be applied also for other inertial locations.\\
As noted in \cite{harle2013survey}, step detection methods can be of four types: peak detection, zero crossing, autocorrelation, and spectral analysis.
The peak detection method detects the maximum peaks of the  accelerometer signal to determine the user step instances \cite{fang2005design,6338075}. In zero crossing detection, the zero crossing of the accelerometer signal is monitored for step detection \cite{6463008,seo2015step}. The cyclic nature of walking leads to strong periodicity in inertial sensor readings. It is possible to extract a cycle from a sequence of sensor data by looking for maxima in the mean-adjusted autocorrelation of the data over a period of time. Autocorrelation-based step detection can be applied to accelerometer \cite{santos2019autocorrelation} or gyroscopes readings \cite{rhudy2018comprehensive}. Steps can also be detected in the frequency domain, for example, by using Fourier \cite{dirican2017step} or wavelet transforms \cite{wang2012real}. \\
Of the methods mentioned above, peak detection is the most popular. These approaches are typically based on the magnitude threshold of the specific force value and the minimum step period.  In this case, a step is defined as the interval between two successive peaks. Moreover, because PDR is a dead-reckoning approach, without external aiding, only short time scenarios can be considered.  Thus, the inertial frame (i-frame) is defined at the user’s starting point. For simplicity, we assume that the body frame (b-frame) coincides with the sensitive axes of the inertial sensors. \\
To formulate the peak detection method, we denote the specific force vector expressed in the body frame, $\mathbf{f}^{b}_{ib}$,  as
\begin{equation} \label{eq:fvec}
    \mathbf{f}^{b}_{ib} = [f_x~ f_y~ f_z]^{T}
\end{equation}
The magnitude of the specific force vector \eqref{eq:fvec} at time $k$ is
\begin{equation} \label{eq:fmag}
    {f}_{mag,k} = \sqrt{f_{x,k}^2+f_{y,k}^2+f_{z,k}^2}
\end{equation}
The mean of the specific force magnitude throughout the trajectory is defined by
\begin{equation} \label{eq:fmagmean}
    \bar{f}_{mag} = \frac{1}{n} \sum^{n}_{k=1}{f}_{mag,k}
\end{equation}
where $n$ is the number of samples. Next, The mean of the specific force magnitude \eqref{eq:fmagmean} is subtracted from the magnitude of the specific force vector \eqref{eq:fmag}
\begin{equation} \label{eq:fmagmean}
    {f}_{m,k} = {f}_{mag,k} - \bar{f}_{mag} 
\end{equation}
Finally, the standard deviation (STD) of the specific force magnitude, after reducing its mean, is calculated by
\begin{equation} \label{eq:fmagstd}
    \sigma_{f}= \left [
    \frac{1}{n} \sum^{n}_{k=1}({f}_{m,k} - \bar{f}_{mag})^2
\right ]^{\frac{1}{2}}    
\end{equation}
 Two parameters are needed to apply a basic peak detection approach:
\begin{enumerate}
    \item The minimum time interval between steps, defined empirically.
    \item The minimum peak height, i.e., the minimum value allowed for a valid peak. It is common practice to use $1.5\sigma_{f}$ as the minimum value. 
\end{enumerate}
We consider the following scenario to illustrate the peak detection algorithm: The user walks holding a smartphone in texting mode, that is,approximately waist-high. The user makes 26 steps to cover a distance of  $25.4$ meters. The specific force was measured by the smartphone accelerometers. Its magnitude, calculated by \eqref{eq:fmag}, and magnitude mean value \eqref{eq:fmagmean} are shown in Figure~\ref{Fig:accmag}. Next, the  mean is subtracted as in \eqref{eq:fmagmean} and the STD is calculated by \eqref{eq:fmagstd}. Finally, the STD is multiplied by a factor of $1.5$ and the resulting value is adopted as the minimum peak height. The minimum time interval between steps was set to 0.3 seconds. Using these two parameters, a simple maxima search is performed to find the user steps. 
Figure~\ref{Fig:steps} shows the specific force magnitude after subtracting its mean. The red circles show maxima values identified as steps. The peak-detection algorithm manged to detect all 26 steps. Note, however, that such approaches are sensitive to user dynamics (walking speed) and sensor location. 
\subsection{Step-length estimation}\label{sec:steplegnth}
\begin{figure}[htbp]
	\centering
    \includegraphics[width=0.5\textwidth]
    {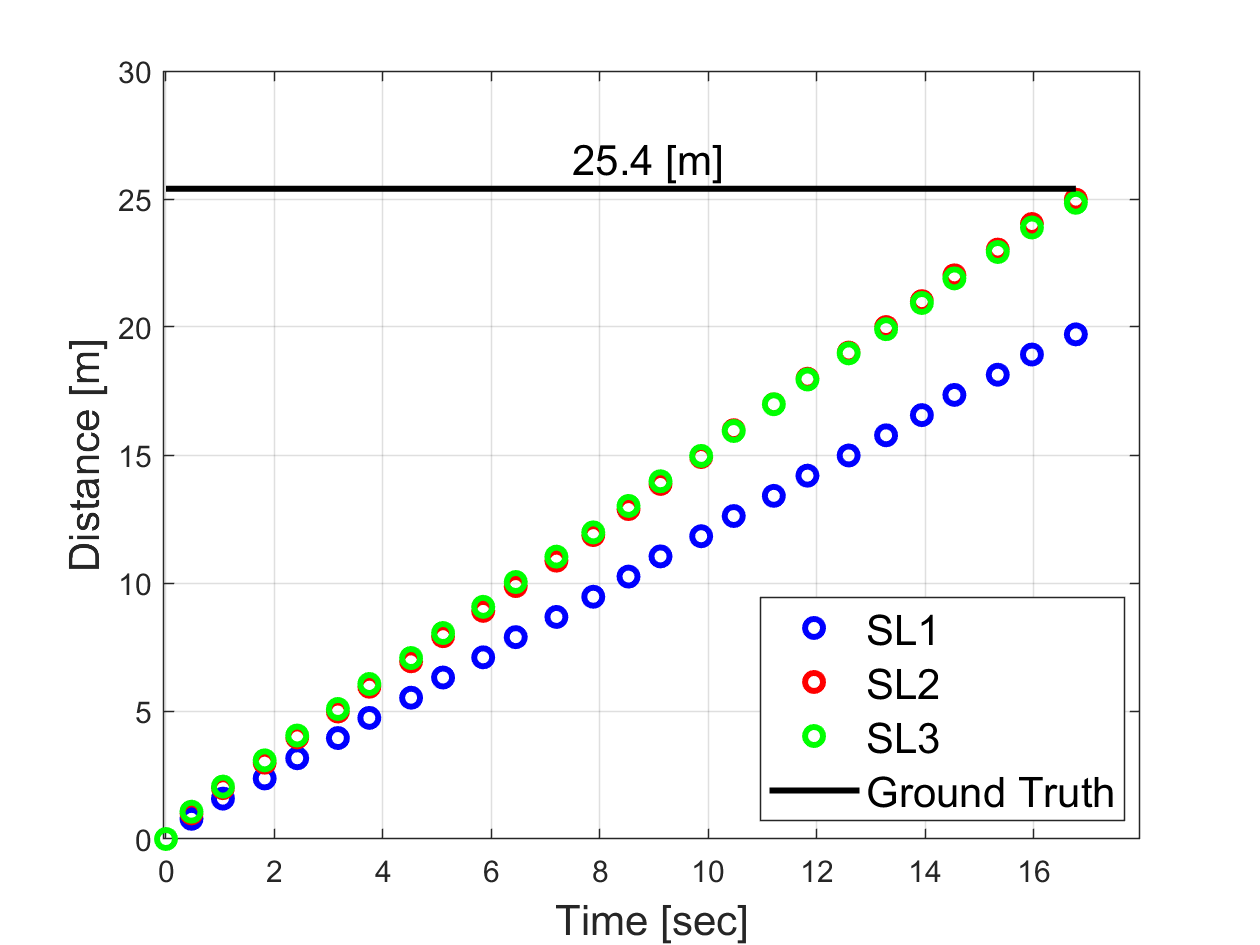}   
    \caption{Estimated user distance of the three step-length approaches. The recording was obtained with a smartphone in texting mode by a pedestrian walking for 25.4 meters.}\label{Fig:SL}
\end{figure}

\begin{figure}[htbp]
	\centering
    \includegraphics[width=0.5\textwidth]
    {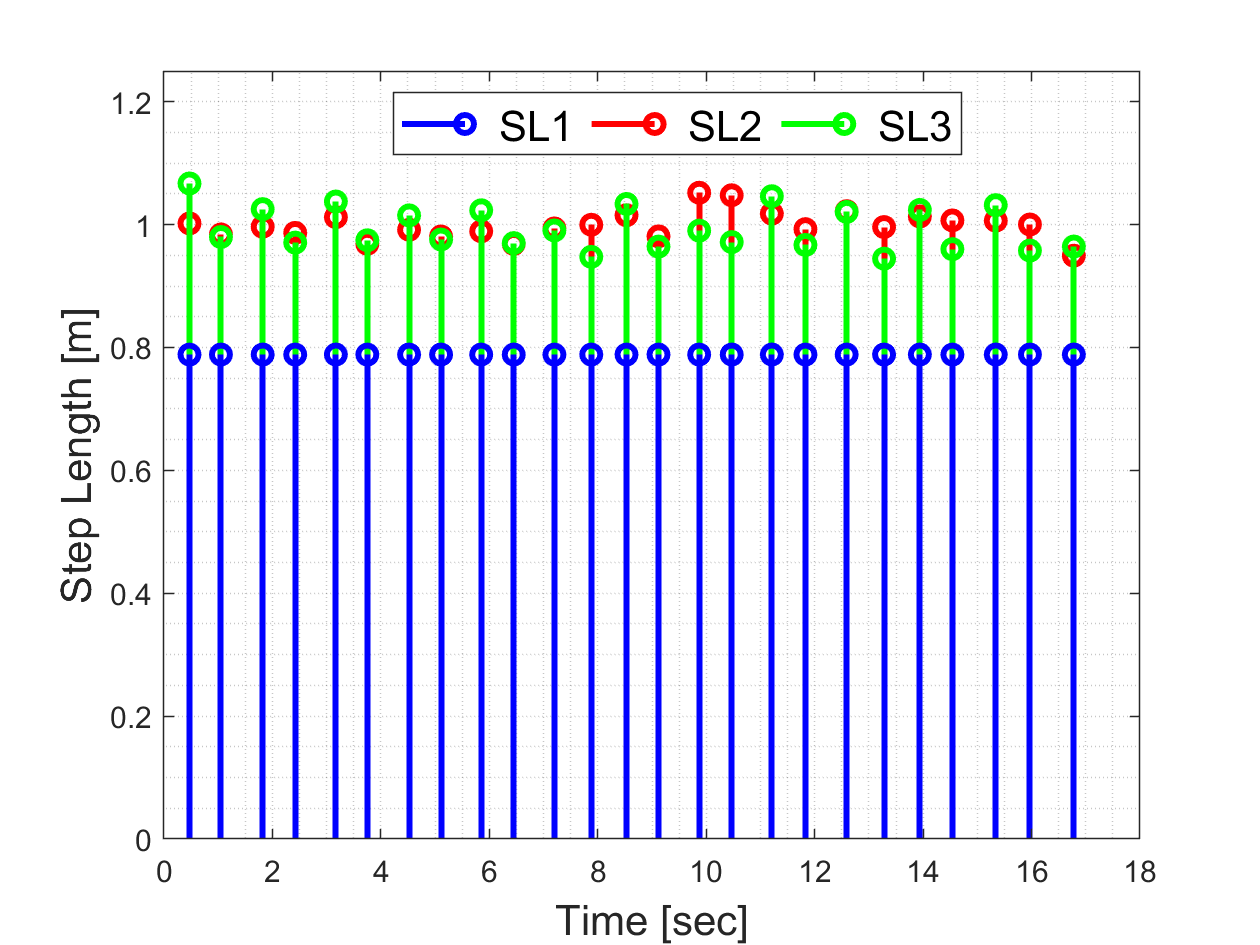}   
    \caption{Lengths of each of the 26 steps made during the trajectory. This recording was taken with a smartphone in texting mode by a pedestrian walking for 25.4 meters.}\label{Fig:step_size}
\end{figure}

\noindent
Pedestrian step-length can be estimated using several approaches, including regression-based, biomechanical models, and empirical relationships. The underlying idea behind these approaches is that a step can be detected based on  accelerometer readings. Therefore, many approaches in the literature attempt to establish a relationship between the specific force exhibited during a step and the length of the step. In a recent study, the use of smartphone inertial sensors for estimating step-length was analyzed and compared with 13 representative model-based step-length estimation models \cite{vezovcnik2018average}. Lately,  \cite{soni2022survey} provided a survey of step-length from various perspectives, including the research method used, the length of the test path, various walking speeds, the location of the sensor device on the user's body, and the accuracy achieved in estimating step-length.\\
We consider three step-length approaches:
\begin{enumerate}
    \item \textbf{SL1}: A constant step-length approach that assumes that all steps during a walk have an equal length and are a function of the user's gender and height \cite{6339316}
    \begin{equation}\label{eq:steplength1}
        s_c = k_c\dot h
    \end{equation}
    where $s_c$ is the constant step-length, $h$ is the user height, and $k_c$ is a gain equal to $0.415$ for men and $0.413$ for women. As it is based solely on user height, this is one of the simplest approaches to estimating step-length. Yet, as a constant value approach, it fails to cope with varying step-lengths.
    \item \textbf{SL2}: The Weinberg \cite{weinberg2002using} is a biomechanical approach, based on the inverted pendulum model. Originally, it relied on the difference between the maximum and minimum vertical acceleration values during a step. Later, it was shown to operate successfully using  the specific force magnitude instead \cite{klein2020stepnet}. Based on the specific force magnitude, the Weinberg approach uses the formula:
      \begin{equation}\label{eq:steplength2}
        s_w = k_w \left (f_{mag,max} - f_{mag,min} \right )^{1/4}
    \end{equation}
    where $k_w$ is the Weinberg gain, $s_w$ is the Weinberg-based step-length, $f_{mag,max}$ is the maximum value of the specific force during the step interval, and $f_{mag,min}$ is the minimum value of the specific force during the step interval. Before using the approach, a calibration procedure, with the user walking a certain distance, should be applied to determine the gain, $s_w$.  Taking the sum operator on both sides of \eqref{eq:steplength2} along the trajectory, allows  determining the gain because the sum of the step-lengths is the known travelled distance. Thus, Weinberg's approach relies on a single empirical gain, which must be calibrated before its use.
    \item  \textbf{SL3}: Adaptive step-length estimation. The step-length of a pedestrian is not constant and varies with walking speed, step frequency, acceleration variance, and other parameters. To increase accuracy in estimating walking distances,  adaptive step-length estimation algorithms have been proposed.  Of these, in \cite{SHIN20111064}, the step-length is a function of step frequency and the variance of the specific force during the step, as given by:
    \begin{equation}\label{eq:steplength3}
        s_a = k_{a,1}\cdot SF + k_{a,2}\cdot \sigma_f + k_{a,3}  
    \end{equation}
    where SF is the step frequency defined as the inverse of the time duration of the step, $\sigma_f$ is the specific force variance \eqref{eq:fmagstd}, and $k_{a,1}-k_{a,3}$ are predefined empirical gains.  As in the Weinberg approach, the determination of the empirical gains requires a calibration procedure.  
\end{enumerate}
Returning to our numeric example from Section~\ref{sec:steplegnth}, we tested the three step-length approaches: SL1-SL3. For SL1, the walking user was a 190 cm-tall male. Calibration procedures with the same user were carried out to obtain the gains of approaches SL2 and SL3.  The estimated distances of the three approaches are shown in Figure~\ref{Fig:SL}. Although the user walked at a constant pace to make sure that his steps were equal, the constant-length approach, SL1, performed worse than the adaptive approaches.  SL2 and SL3 obtained a distance error of less than $2\%$ of the travelled distance. \\
Figure~\ref{Fig:step_size} shows the the lengths of each of the 26 steps made during the trajectory. In SL1, a constant step-length of 79cm was obtained. In both SL2 and SL3, the average step-length was 99cm, reflecting the actual step-length.\\ 
Step-length approaches are sensitive to user characteristics (like height and weight), inertial sensors location, and walking speeds.
\subsection{Heading and walking direction}
\begin{figure}[htbp]
	\centering
    \includegraphics[width=0.4\textwidth]
    {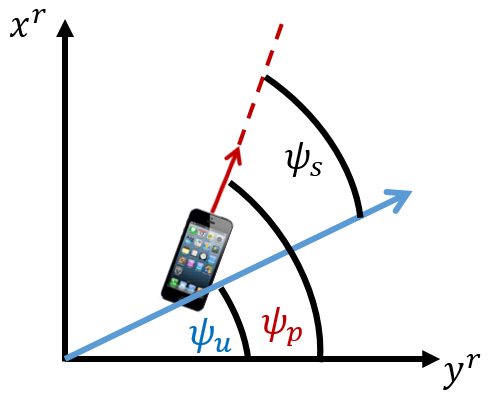}   
    \caption{Heading geometry defined relative to the starting point of the trajectory. The figure shows the offset angle between the smartphone inertial sensors (red) and the user's walking direction (blue).  }\label{Fig:heading_angles}
\end{figure}

\begin{figure}[htbp]
	\centering
    \includegraphics[width=0.45\textwidth]
    {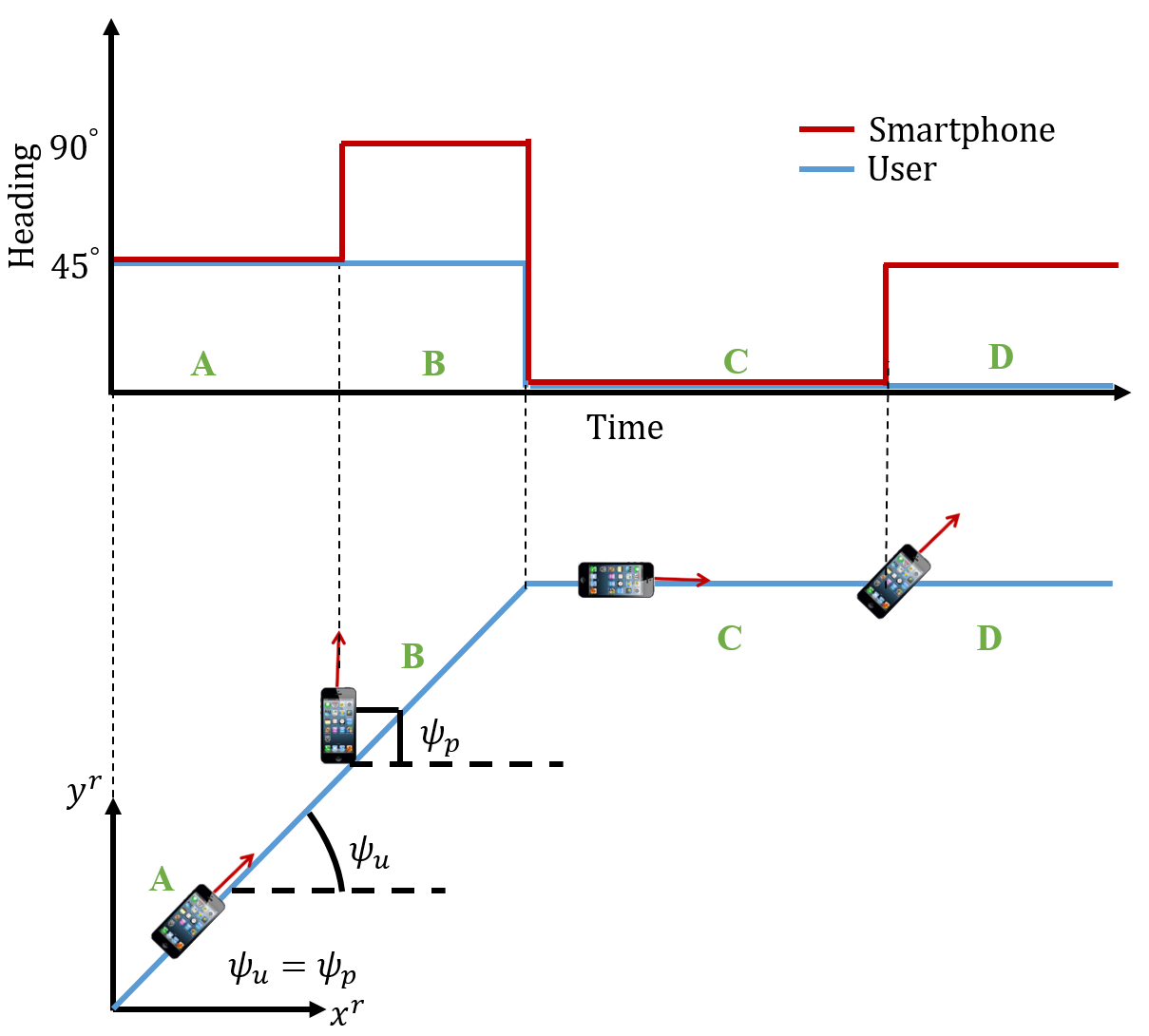}   
    \caption{A user walks holding the smartphone at different angles relative to the walking direction. The lower part of the figure shows the user trajectory and the upper part the heading angles of the user and smartphone across the trajectory.}\label{Fig:heading_traj}
\end{figure}

\noindent
To estimate the pedestrian's trajectory, it is necessary to estimate the  walking direction and heading. In some cases, the direction of walking does not coincide with the direction of the inertial sensor's sensitive axis.  For example, when using the smartphone, the user may point it with an offset to the  walking direction, as illustrated in Figure~\ref{Fig:heading_angles}. 
The figure shows the heading geometry defined relative to the starting point of the trajectory. From the geometry we define:
\begin{equation}\label{eq:heading_geo}
    \psi_u = \psi_p - \psi_s
\end{equation}
where $\psi_u$ is the user's heading (walking direction), $\psi_p$ is the heading angle of the smartphone (or any other inertial sensor), and $\psi_s$, is the sliding (offset) angle between the smartphone and the user. 
Equation \eqref{eq:heading_geo} represents a general walking scenario. To better illustrate the difference, consider the scenario shown in Figure~\ref{Fig:heading_traj}, where a user holding a smartphone starts walking while the smartphone is aligned to the walking direction (segment A). During the walking, the user maintains the same direction but changes the smartphone direction by $45^{\circ}$, as shown in segment B. Next, simultaneously, the user changes  walking direction by $45^{\circ}$ and  smartphone direction by $90^{\circ}$,  aligning the two, as shown in segment C. Finally, in segment D, the user rotates the phone by $45^{\circ}$ relative to the  walking direction. The user trajectory and both smartphone and user heading angles are illustrated in Figure~\ref{Fig:heading_traj}.  \\  
Two spacial cases may occur: (a) the inertial sensors are held in the same direction as the walking direction, satisfying $\psi_s(t)=0, \forall t$, for example, a  user walking and holding the smartphone in texting mode; (b) the inertial sensors are rigidly mounted to the user, resulting in a constant offset angle $\psi_s(t)=\psi_{s,c}, \forall t$. \\
One of the most commonly used methods for estimating the heading of smartphones, $\psi_p$, is based on fusing data from accelerometers, magnetometers, and gyroscopes in an attitude and heading reference system (AHRS) framework. There are many types of AHRS approaches, including the Kalman filter~\cite{choukroun2006novel,lee2012estimation}, nonlinear estimation~\cite{sabatini2011kalman, munguia2011attitude}, complementary filtering~\cite{mahony2008nonlinear,fourati2010nonlinear,madgwick2011estimation}, and hybrid-learning approaches~\cite{vertzberger2021attitude, vertzberger2021attitudec, vertzberger2022adaptive}. Regardless of the AHRS algorithm choice, they all rely on the  kinematic equations of the gyroscope. Following ~\cite{farrell2008aided}, we provide a simplified description of these equations, assuming initial zero roll and pitch angles, and that the x-axis of the smartphone is aligned with the walking direction, resulting in an initial zero heading. This corresponds to an initial quaternion of $\textbf{q}_0\in\mathbb{R}^4$:
\begin{equation}\label{eq:q0}
    \textbf{q}_0 = 1 + 0\textbf{i} + 0\textbf{j} + 0\textbf{k} 
\end{equation}
where 1, $\textbf{i},\textbf{j},\textbf{k}$ are the quaternion basis elements.\\
Given the gyroscope's measurement, $[\omega_x~\omega_y~\omega_z]^{T}$, the kinematic equation for the quaternion (its rate of change in time) is
\begin{equation}\label{eq:qdot}
    \dot{q} = \frac{1}{2} \left [ \begin{array}{ccc}
        -q_2 & -q_3 & -q_4 \\
        q_1 & q_4 & -q_3 \\
        -q_4 & q_1 & q_2 \\
        q_3 & -q_2 & -q_1 
    \end{array} \right ] \left [ \begin{array}{c}
         \omega_x \\
         \omega_y \\
         \omega_z 
    \end{array} \right ]
\end{equation}
Given $\dot{q}$ and the previous quaternion, the current quaternion can be calculated to construct the transformation matrix from the reference  frame to the body frame:
{ \footnotesize
\begin{equation}\label{eq:t_rb}
    \textbf{T}^{b}_{r} = \left [ \begin{array}{ccc}
        q^2_1+q^2_2-q^2_3-q^2_4 & 2(q_2q_3-q_1q_4) & 2(q_1q_3+q_2q_4) \\
        2(q_2q_3+q_1q_4) & q^2_1-q^2_2+q^2_3-q^2_4 & 2(q_3q_4-q_1q_2) \\
        2(q_2q_4-q_1q_3) & 2(q_1q_2+q_3q_4) & q^2_1-q^2_2-q^2_3+q^2_4
    \end{array} \right]
\end{equation}
}
Using \eqref{eq:t_rb}, the smartphone heading is calculated by
\begin{equation}\label{eq:psi_ahrs}
    \psi_p = atan2 \left ( 2(q_2q_3-q_1q_4), 1-2(q^2_3+q^2_4)   \right)
\end{equation}
As noted, using \eqref{eq:psi_ahrs} for the estimation of the user's walking direction is valid only when the smartphone is aligned with the user's walking path so that $\psi_s=0$ and thus $\psi_u=\psi_p$. If this is not the case, other approaches to estimate  the user's walking direction are needed. 
The walking direction may be extracted from acceleration measurements using a kinematic “rolling-foot” model~\cite{leonardo2019determination}, by principle component analysis of the accelerometer readings~\cite{deng2015heading}, or using gravity-based ~\cite{manos2018gravityc,thio2021relative} and deep learning approaches~\cite{manos2022walking, wang2019pedestrian}.\\
Here, we describe a simple yet efficient gravity-based approach for estimating the user's walking direction. First, the gravity direction vector is obtained and the angular velocity is projected to obtain its vertical component, which is integrated to find the walking direction.
As stated in \cite{manos2019gravity}, to reduce or eliminate the large temporal variations typically associated with the acceleration of a pedestrian in motion, it is necessary to apply low-pass filtering to each axis of the force vector. Then, the gravity vector direction is given by
\begin{equation}\label{eq:grav_dir}
    \boldsymbol{\gamma}_g = \frac{-\textbf{f}_{LPF}}{||\textbf{f}_{LPF}||}
\end{equation}
where $\mathbf{\gamma}_g$ is the gravity direction vector and $\textbf{f}_{LPF}$ is the filtered specific force vector in the sensor frame.\\
Denoting the vertical projection of the gyroscope reading by $\omega_v$ and the measured angular velocity by  $\boldsymbol{\omega}$, it can be calculated at any instance of time by
\begin{equation}\label{eq:omega_vec}
    \omega_v = \boldsymbol{\gamma}_g^{T} \boldsymbol{\omega}
\end{equation}
As $\omega_v$ measures the pedestrian’s turning rate in the horizontal plane, an approximation for the change in the walking direction during the time interval $\Delta t$ is 
\begin{equation}\label{eq:delta_psi}
    \Delta\psi_u = \omega_v\Delta t
\end{equation}
The \eqref{eq:grav_dir}-\eqref{eq:delta_psi} approach for finding the change in the user's walking direction is valid assuming that the angular velocity is constant during the time interval $\Delta t$, which holds for most gyroscopes sampling at a high-rate. Notice further that to filter out the user acceleration using a low-pass filter, the user should be walking. In transient walking motion, as at the beginning or end of walking, we expect a degradation in filtering quality, leading to large heading errors.
\subsection{Two-dimensional positioning}
\noindent
Given the estimated step size, $s$, and the user's walking direction angle, $\psi_u$, the two-dimensional PDR user position vector is:
\begin{equation}\label{eq:pdr_eom}
\left [\begin{array}{c}
     x_k  \\
     y_k
\end{array} \right ] =
 \left [\begin{array}{c}
     x_{k-1} + s_k\cos(\psi_{u,k})  \\
     y_{k-1} + s_k\sin(\psi_{u,k})
\end{array} \right ]   
\end{equation}
where $x$ and $y$ are the two-dimensional position components and $k$ is the current epoch. \\
Algorithm~\ref{algo1} summarizes the two-dimensional model-based PDR method.
\begin{algorithm}[h!]
\caption{Two-dimensional model-based PDR}\label{algo1}
\KwIn{$\mathbf{f}^b,\boldsymbol{\omega}^b,x_{k-1},y_{k-1},s_k,\psi_{u,k},\psi_{s,k},\psi_{p,k}$}
\textbf{Step detection:} $\mathbf{f}^b$ and \eqref{eq:fvec}-\eqref{eq:fmagstd} \;
\textbf{step-length estimation:} $\mathbf{f}^b$ and one approach out of \eqref{eq:steplength1}, \eqref{eq:steplength2}, or \eqref{eq:steplength3} \;
\textbf{Heading determination:} 
\eIf{$\psi_{s,k}=0$}{
   $\boldsymbol{\omega}^b$ and \eqref{eq:qdot}-\eqref{eq:psi_ahrs} such that $\psi_{u,k}=\psi_{p,k}$ \;
   }{
   $\mathbf{f}^b$ and \eqref{eq:grav_dir}-\eqref{eq:delta_psi}\;
  }
  \textbf{Two-dimensional positioning:} $x_{k-1},y_{k-1},s_k,\psi_{u,k}$ and \eqref{eq:pdr_eom} \;  
\KwOut{$x_k, y_k$}
\end{algorithm}
\subsection{Three-dimensional PDR}
\begin{figure}[htbp]
	\centering
    \includegraphics[width=0.5\textwidth]
    {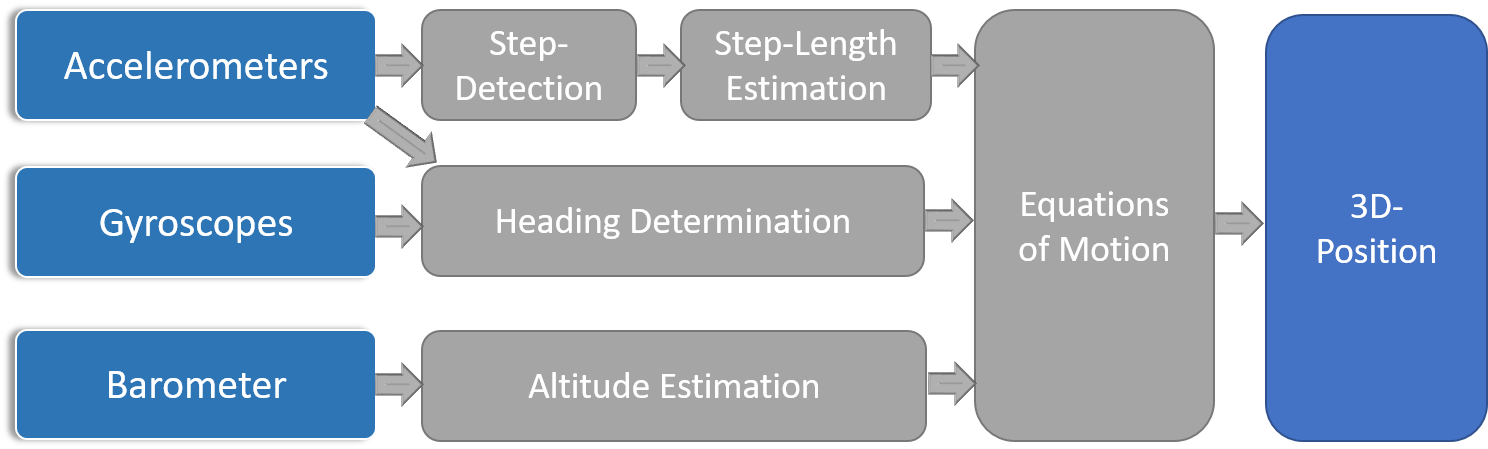}   
    \caption{Model-based 3D PDR using a barometer for altitude estimation.}\label{Fig:3dbaro}
\end{figure}

Most PDR approaches assume that the user walks in the same plane, without vertical movement. Yet, in typical walking environments vertical movement is required, for example, when using staircases, elevators, or
escalators. A typical solution for solving the altitude problem is to use a barometer to determine the change in altitude, as applied in \cite{6393106,6236898}. Given a barometer, the 3D PDR algorithm, for each sensor role, is presented in Figure~\ref{Fig:3dbaro}.
Yet, as pointed out in \cite{boim2021height}, not all smartphones or wearable devices have a barometer. In these cases, the altitude can be estimated using accelerometer measurements. In \cite{itzik2019step}, a modified Weinberg approach was suggested to estimate the user's steps during movement on a staircase and a calibration phase is required to estimate the Weinberg gain.
A later study, \cite{boim2021height}, proposed estimating the change in height by identifying peaks in the motion of the user during movement on a staircase in the course of which the user is changing position. This approach does not require prior calibration.
\section{Shoe-Mounted INS}\label{sec:ShoeMountedIMU}
\begin{figure}[htbp]
	\centering
    \includegraphics[width=0.5\textwidth]
    {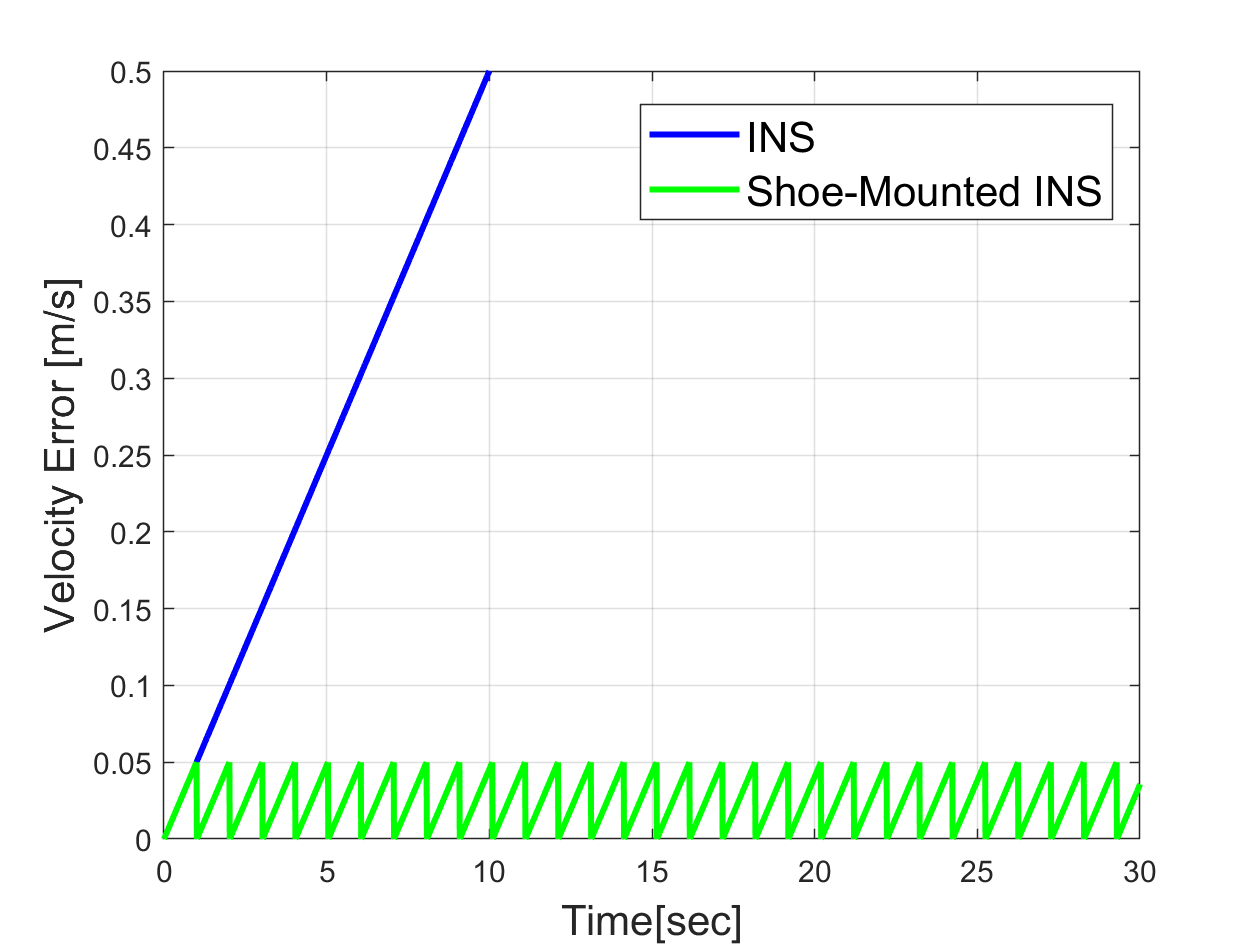}   
    \caption{Illustration of the INS velocity solution drift (without aiding) and the shoe-mounted INS with information aiding during zero velocity instances.}\label{Fig:SM_velerror}
\end{figure}

\begin{figure}[htbp]
	\centering
    \includegraphics[width=0.4\textwidth]
    {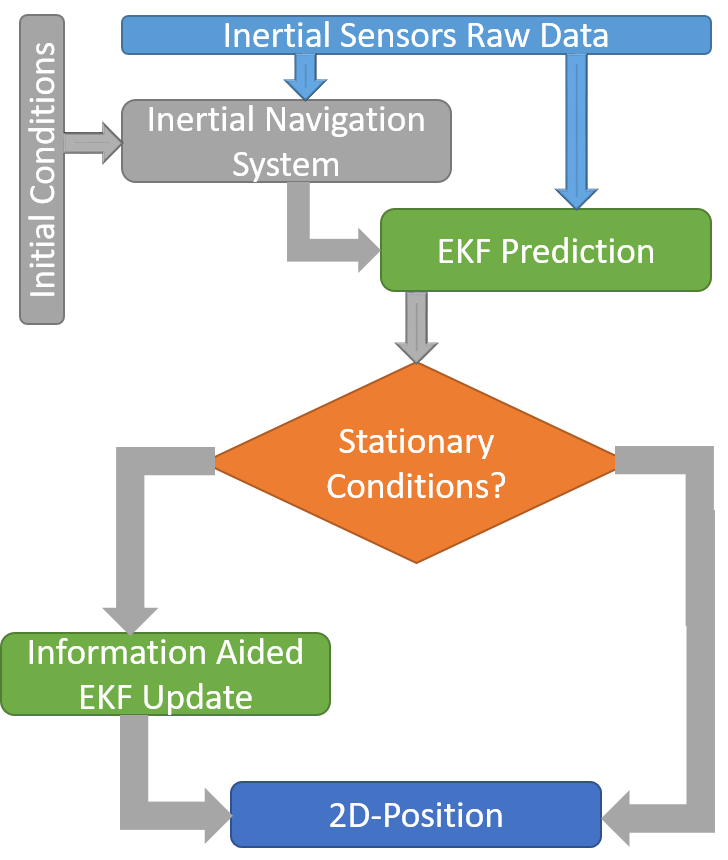}   
    \caption{Shoe-mounted INS framework.}\label{Fig:showins}
\end{figure}

\noindent
In SM-INS (or foot-mounted) devices, the inertial sensors are rigidly mounted on a shoe\cite{Foxlin2005, nilsson2012foot}, so that zero velocity events are identified and information aiding can be applied to mitigate the inertial drift \cite{engelsman2023information}. Therefore, instead of experiencing a velocity drift caused by a strapdown INS mechanism, a shoe-mounted INS produces a saw-tooth like velocity error behaviour. Figure~\ref{Fig:SM_velerror} illustrates this assuming ideal initial conditions and only accelerometer bias. 
In this example, every one-second stationary condition is identified and information aiding is applied to nullify velocity drift.  In practice, the same velocity error behavior is observed even when taking into account all accelerometer and gyroscope errors. \\
The shoe-mounted INS framework, using an extended Kalman filter (EKF), is illustrated in Figure~\ref{Fig:showins}. Note that any other nonlinear filter can be applied instead.
The INS navigation solution is produced in accordance with the inertial conditions and inertial readings. The inertial measurements, together with the INS solution, are used in the EKF prediction phase. Next, an algorithm to detect zero velocity instances is applied. If such instances are detected, information aiding is used in the EKF update phase to produce a corrected navigation solution. Otherwise, the INS solution is not corrected by the filter.
We examine each component of the shoe-mounted INS in detail below. 
\subsection{Inertial Navigation}\label{sec:INS}
\begin{figure}[htbp]
	\centering
    \includegraphics[width=0.4\textwidth]
    {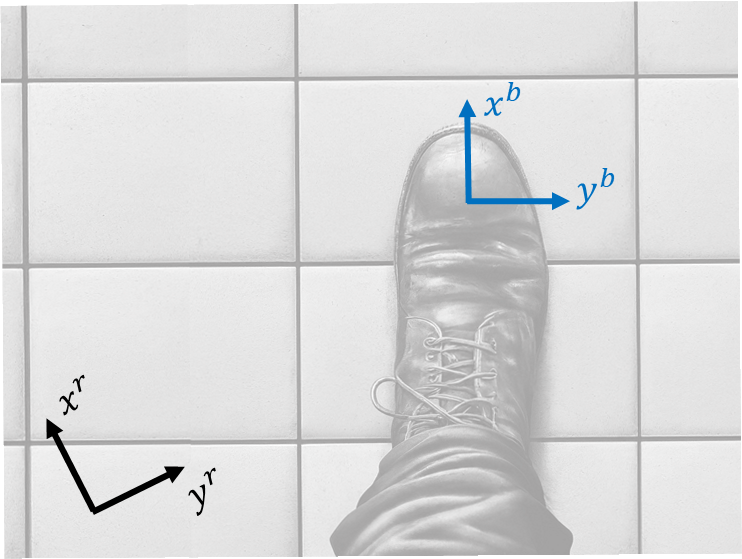}   
    \caption{Reference frames associated with a typical shoe-mounted INS scenario.}\label{Fig:shoe}
\end{figure}

\noindent
The INS equations of motion are a set of first order differential equations. Given initial conditions and inertial measurements, they can be solved to give the navigation solution, namely the position, velocity, and orientation. These equations are valid for any platform,  regardless of its operating environment \cite{titterton2004strapdown}. In shoe-mounted INS, however, some simplification is applied, reducing the complexity of the INS equations of motion.  Generally, low-performance inertial sensors are used in shoe-mounted INS, therefore, angular velocity vector of the earth can be neglected. The transport rate is also neglected because the user velocity is relatively low. Last, it is assumed that the body frame coincides with the inertial sensor frame located on the shoe, as illustrated in Figure~\ref{Fig:shoe}. 
Taking those assumptions into account, the continuous form of the INS equation of motion is:
\begin{equation}\label{eq:ins_veldot}
    \dot{\textbf{v}}^{r} = \textbf{T}^{b}_{r}\textbf{f}^b_{ib} + \textbf{g}^{r}
\end{equation}
where $\dot{\textbf{v}}^{r}$ is the velocity vector expressed in the reference frame,  $\textbf{f}^b_{ib}$ is the specific force vector expressed in the body frame, $\textbf{T}^{b}_{r}$ is the reference to body transformation matrix defined in \eqref{eq:t_rb}, and $\textbf{g}^{r}$ is the local gravity vector defined by:
\begin{equation}\label{eq:grav_00g}
    \textbf{g}^{r} = \left [ 0,~0,~g \right ]^{T}
\end{equation}
where $g$ is the local gravity, assumed constant during the motion.\\
The position rate of change, $\dot{\textbf{p}}^{r}$,  is the integration on the velocity vector:
\begin{equation}\label{eq:ins_posdot}
    \dot{\textbf{p}}^{r} =  \textbf{v}^{r}
\end{equation}
Last, the transformation matrix rate of change is
\begin{equation}\label{eq:ins_Tdot}
    \dot{\textbf{T}}^{b}_{r} = \textbf{T}^{b}_{r}\boldsymbol{\omega}^b_{ib}
\end{equation}
where ${\omega}^b_{ib}$ is the measured gyroscope angular velocity vector.\\
Given inertial conditions, the INS equations of motion, \eqref{eq:ins_veldot}, \eqref{eq:ins_posdot}, and \eqref{eq:ins_Tdot} are integrated to obtain the navigation solution. The choice of the numeric integration algorithms is a trade-off between precision and processing capability. For a detailed discussion of this topic, refer to~\cite{titterton2004strapdown,Groves2013}. In SM-INS, the Pad$\Acute{\text{e}}$ approximation is regularly used to propagate the transformation matrix \cite{fischer2012tutorial,jimenez2010indoor}:
\begin{equation}\label{eq:pade}
    \textbf{T}^{b}_{r,k} = \textbf{T}^{b}_{r,k-1}
    \left ( 2\textbf{I}_3+\boldsymbol{\Omega}_k\Delta t
    \right )
     \left ( 2\textbf{I}_3-\boldsymbol{\Omega}_k\Delta t
    \right )^{-1}
\end{equation}
where $\Delta t$ is the time-step, $k$ is the current epoch, and $\boldsymbol{\Omega}_k$ is the skew symmetric of the measured angular velocity
\begin{equation}\label{eq:Omega}
 \boldsymbol{\Omega}_k = 
 \left [ \begin{array}{ccc}
      0    &  -\omega_{z,k} & \omega_{y,k} \\
      \omega_{z,k}& 0  &  -\omega_{x,k} \\
      -\omega_{y,k} & \omega_{x,k} & 0
 \end{array} \right ]
\end{equation}
The velocity is updated using the Euler method:
\begin{equation}\label{eq:ins_velupdate}
    \textbf{v}^{r}_{k} = \textbf{v}^{r}_{k-1}+\Delta t\dot{\textbf{v}}^{r}
\end{equation}
where $\dot{\textbf{v}}^{r}$ is defined in \eqref{eq:ins_veldot}. Similarly, the position update is given by:
\begin{equation}\label{eq:ins_posupdate}
    \textbf{p}^{r}_{k} = \textbf{p}^{r}_{k-1}+\Delta t\dot{\textbf{p}}^{r}
\end{equation}
where $\dot{\textbf{p}}^{r}$ is defined in \eqref{eq:ins_posdot}.
\subsection{Navigation Filter}\label{sec:NavFilter}
\noindent
The nonlinear nature of the INS equations of motion necessitates a nonlinear filter when INS is fused with external sensors or information, as in SM-INS. Before designing the filter, we define the inertial sensor measurement error model as:
\begin{align}\label{eq:fw}
\Tilde{\boldsymbol{f}}_{ib}^b &= \boldsymbol{f}_{ib}^b + \textbf{b}_a + \textbf{w}_a \\
\Tilde{\boldsymbol{\omega}}_{ib}^b &= \boldsymbol{\omega}_{ib}^b + \textbf{b}_g + \textbf{w}_g 
\end{align}
where $\textbf{w}_a$ and $\textbf{w}_g$ are the zero mean white Gaussian noise of the accelerometer and gyroscope measurement, respectively, and the sensor biases are modeled as random walk processes
\begin{align} \label{err_b_g}
\dot{\textbf{b}}_a  & =  \textbf{w}_{ab} \\
\dot{\textbf{b}}_g  & =  \textbf{w}_{gb} 
\end{align}
where $\textbf{b}_a$ is the accelerometer bias,  $\textbf{b}_g$ is the gyroscope bias, and $\textbf{w}_{ab}$ and $\textbf{w}_{gb}$ are the zero mean white Gaussian noise of the accelerometer and gyroscope bias modeling, respectively.\\
An error-state EKF implementation is used with an  error-state vector
\begin{equation}\label{eq:iInsErrorState}
\delta \mathbf{x} = \left[ \begin{array}{ccccc} 
\delta\mathbf{p}^{r} & \delta\mathbf{v}^{r} & \mathbf{\epsilon}^{r} & \mathbf{b}_{a} & \mathbf{b}_{g} \end{array} \right]^{\operatorname{T}} \in \mathbb{R}^{15}
\end{equation}
where $\delta\mathbf{p}^{r}$ is the position error-state vector, $\delta\mathbf{v}^{r}$ is the velocity error-state vector, and $\boldsymbol{\epsilon}^{r}$ is misalignment error-state. \\
The linearized error-state model is \cite{Groves2013}
\begin{equation}\label{eq_errorModel}
\delta\dot{\mathbf{x}} = \mathbf{F}\delta\mathbf{x}  + \mathbf{G}\mathbf{w}
\end{equation}
where $\mathbf{F}$ is the system matrix, $\mathbf{G}$ is the shaping matrix, and $\delta\mathbf{w}$ is the noise vector.
The residuals of the accelerometers and gyros  are modeled as random walk processes although any other suitable models may be used instead, such as the first-order Gauss-Markov process. The system matrix is given by
\begin{equation}\label{eq_Fmat}
\mathbf{F} = \left[ \begin{array}{ccccc}
 \mathbf{0}_{3\times3} & \mathbf{I}_{3} &\mathbf{0}_{3 \times 3} & \mathbf{0}_{3\times 3} & \mathbf{0}_{3\times 3} \\
\mathbf{F}_{vp} & \mathbf{0}_{3\times3} & \mathbf{F}_{v\epsilon} & \mathbf{T}_{b}^{n} & \mathbf{0}_{3\times 3} \\
\mathbf{0}_{3\times3} & \mathbf{0}_{3\times3} & \mathbf{0}_{3\times3} & \mathbf{0}_{3\times 3} & \mathbf{T}_{b}^{n} \\
\mathbf{0}_{3\times3} & \mathbf{0}_{3\times 3} & \mathbf{0}_{3\times 3} & \mathbf{0}_{3\times 3} & \mathbf{0}_{3\times 3} \\
\mathbf{0}_{3\times3} & \mathbf{0}_{3\times 3} & \mathbf{0}_{3\times 3} & \mathbf{0}_{3\times 3} & \mathbf{0}_{3\times 3} 
\end{array}
\right]. 
\end{equation}
where $F_{ij}$ are $3\times 3$  submatrices obtained from the linearization of the nonlinear equation of motion (more details on the internalization process can be found in navigation textbooks such as \cite{farrell2008aided, Groves2013}). The shaping matrix is given by
\begin{equation}\label{eq_Gmat}
\mathbf{G} = \left[ \begin{array}{cccc}
\mathbf{0}_{3\times3} & \mathbf{0}_{3\times 3} & \mathbf{0}_{3\times 3} & \mathbf{0}_{3\times 3} \\
\mathbf{T}_{b}^{n} & \mathbf{0}_{3\times 3} & \mathbf{0}_{3\times 3} & \mathbf{0}_{3\times 3} \\
\mathbf{0}_{3\times3} & \mathbf{T}_{b}^{n} & \mathbf{0}_{3\times 3} & \mathbf{0}_{3\times 3} \\
\mathbf{0}_{3\times3} & \mathbf{0}_{3\times 3} & \mathbf{I}_{3} & \mathbf{0}_{3\times 3} \\ 
\mathbf{0}_{3\times3} & \mathbf{0}_{3\times 3} &  \mathbf{0}_{3\times 3} &\mathbf{I}_{3} 
\end{array}
\right] 
\end{equation}
and the noise vector is
\begin{equation}\label{eq_NoiseVec}
\mathbf{w} = \left[ \begin{array}{cccc} 
\mathbf{w}_a & \mathbf{w}_g & \mathbf{w}_{ab} & \mathbf{w}_{ab} \end{array} \right]^{\operatorname{T}}
\end{equation}
The implementation algorithm for the prediction phase of the EKF error-state closed loop is
\begin{eqnarray}\label{eq:EKF1}
\delta\hat{\mathbf{x}}^{-}_{k} & = & 0 \\
\mathbf{P}^{-}_{k} & = & \mathbf{\Phi}_{k-1}\mathbf{P}^{+}_{k-1}\mathbf{\Phi}^{\operatorname{T}}_{k-1}+\mathbf{Q}_{k-1} \label{eq:EKFend1}
\end{eqnarray}
where  $\delta\mathbf{x}^{-}_{k}$ is the \textit{a priori} estimate of the error-state, $\mathbf{P}^{-}_{k}$ is the covariance of the \textit{a priori estimation error}, $\mathbf{\Phi}_{k}$ is the state transition matrix, and $\mathbf{Q}_{k}$ is the process noise covariance assumed to be constant throughout the trajectory.\\
The EKF update phase is:
\begin{eqnarray}\label{eq:EKF2}
\delta\hat{\mathbf{x}}^{+}_{k} & = & \mathbf{K}_{k}\delta\mathbf{z}_{k} \\
\mathbf{P}^{+}_{k} & = & [\mathbf{I}-\mathbf{K}_{k}\mathbf{H}_{k}]\mathbf{P}^{-}_{k} \\
\mathbf{K}_{k} & = & \mathbf{P}^{-}_{k}\mathbf{H}^{\operatorname{T}}_{k}[\mathbf{H}_{k}\mathbf{P}^{-}_{k}\mathbf{H}^{\operatorname{T}}_{k}+\mathbf{R}_{k}]^{-1} \label{eq:EKFend2}
\end{eqnarray}
where $\delta\mathbf{x}^{+}_{k}$ is the \textit{a posteriori} estimate of the error-state,   $\mathbf{P}^{+}_{k}$ is the covariance of the \textit{a posteriori} estimation error, $\delta \mathbf{z}_{k}$ is the measurement residual vector,  $\mathbf{K}_{k}$ is the Kalman gain, $\mathbf{R}_{k}$ is the measurement noise covariance assumed to be constant for all samples, and  $\mathbf{H}_{k}$ is the measurement matrix.
\subsection{Zero Velocity Detectors}
\noindent
The function of zero velocity detectors (ZVD) in SM-INS is to determine whether information aiding can be applied at the filter update stage \eqref{eq:EKF2}-\eqref{eq:EKFend2}. In a recent review of ZVD, Wahlström and Skog \cite{9174869} recount the history of foot-mounted inertial navigation, characterize the main sources of error, and analyze current approaches to robust ZVD. These methods include heuristic approaches, adaptive thresholding, gait cycle segmentation, other model-based approaches, as well as data-driven methods. \\
There exist several heuristic ZVD that rely on different inertial features, such as the acceleration variance or magnitude, the angular rate energy detector, and the stance hypothesis optimal detection (SHOE) detector. As pointed out in \cite{5523938}, ZVD can be formulated as a likelihood ratio check. To this end, let $\mathcal{H}_0$ denote a hypothesis that the inertial sensor unit is moving and $\mathcal{H}_1$ a hypotheses that the inertial sensor unit is stationary. Note that stationary conditions occur during walking in the \textit{stance} phase or while standing still. The likelihood-ratio test based on inertial measurements, $\textbf{z}_j$, decides on hypothesis $\mathcal{H}_1$ if and only if
\begin{equation}\label{eq:logliklihood}
    L(\textbf{z}_j) = \frac{p\left ( \textbf{z}_j| \mathcal{H}_1\right)}
                           {p\left ( \textbf{z}_j| \mathcal{H}_0\right)}>\gamma   
\end{equation}
where $\gamma$ is some user-defined threshold and  $\textbf{z}_j) = {m_n}^{j+w_f}_{j-w_b}$ with $w_f$ and $w_b$ as the forward and backward window length applied to the required set of measurements. \\
A multi-condition bringing together both accelerometer and gyroscope measurements was proposed in \cite{jimenez2010indoor}. It includes three conditions for declaring a foot to be stationary:
\begin{enumerate}
    \item \textbf{Acceleration magnitude}. Although referred to as acceleration, this is actually a condition concerning the \textit{specific force} magnitude defined by:
    \begin{equation}\label{eq:C1}
        \text{C1} = \left \{ \begin{array}{cc}
             1 & \gamma_{fmag,min}   <  \textbf{f}_{mag,k} < \gamma_{fmag,max}\\
             0 & \text{otherwise}  
        \end{array} \right .
    \end{equation}
    where $\textbf{f}_{mag,k}$ is defined in \eqref{eq:fmag}, and $\gamma_{fmag,min}$ and $\gamma_{fmag,max}$ are the minimum and maximum threshold values, respectively.
    \item \textbf{Local acceleration variance}. Defines the foot activity as
    \begin{equation}
        \sigma^2_f = \frac{1}{2w+1}\sum^{j=k+w}_{j=k-w} (\textbf{f}_{mag,k} - \bar{\textbf{f}}_{mag})^2
    \end{equation}
    where $w$ is the window size and $\bar{\textbf{f}}_{mag}$ is defined in \eqref{eq:fmagmean}. 
    The second condition is satisfied when
        \begin{equation}\label{eq:C2}
        \text{C2} = \left \{ \begin{array}{cc}
             1 & \sigma^2_f>\gamma_{\sigma_f}\\
             0 & \text{otherwise}  
        \end{array} \right .
    \end{equation}
    where $\gamma_{\sigma_f}$ is the local acceleration variance value.
    \item \textbf{Angular velocity magnitude}. This condition requires that the angular velocity magnitude
    \begin{equation}\label{eq:omegamag}
        \boldsymbol{\omega}_{mag,k} = \sqrt{\omega^2_{x,k}+\omega^2_{y,k}+\omega^2_{z,k}}
    \end{equation}
    must be below a given threshold $\gamma_{\sigma_{\omega}}$
     \begin{equation}\label{eq:C3}
        \text{C3} = \left \{ \begin{array}{cc}
             1 & \sigma^2_f>\gamma_{\sigma_{\omega}}\\
             0 & \text{otherwise}  
        \end{array} \right .
    \end{equation}
\end{enumerate}
Detection of a stationary foot requires that all three logical conditions be satisfied simultaneously, so a logical "AND" is applied, and the result is filtered out with a neighboring window median filter. \\
Rather than use heuristic or other model-based approaches for ZVD,  several papers have recently explored the possibility of using machine learning approaches to identify zero velocity instances. For example, \cite{park2016stance} used support vector machines and  \cite{wagstaff2019robust} a long
short-term memory (LSTM) neural network.
\subsection{Information Aiding}
\noindent
Once a zero velocity instance is detected, information aiding can be applied in the EKF update phase \eqref{eq:EKF2}-\eqref{eq:EKFend2}. Mostly, two types of information aiding are used in SM-INS: (1) zero velocity update (ZVU) and (2) zero angular rate (ZAR) \cite{engelsman2023information}.
\begin{enumerate}
    \item \textbf{Zero velocity update} \\
    The principle of this aiding states that while in stationary conditions, the velocity vector in the reference frame is zero. The ZUV measurement residual is given by:
    \begin{align} \label{eq:zzvu}
    \delta \mathbf{z}_{_{\textbf{ZVU}}} &= \textbf{v}_{_{\text{INS}}}^r - \textbf{0}_{3 \times 1} = \textbf{H}_{_{\textbf{ZVU}}} \delta \mathbf{x} + \boldsymbol{\nu}_{_{\textbf{ZVU}}}
    \end{align}
    where $\textbf{v}_{_{\text{INS}}}^r$ is the calculated INS velocity vector \eqref{eq:ins_velupdate} and $\boldsymbol{\nu}_{_{\textbf{ZVU}}}$ is zero mean white Gaussian measurement noise. The corresponding measurement matrix is defined by:
    \begin{equation}
    \textbf{H}_{_{\textbf{ZVU}}} = \begin{bmatrix}
    \textbf{0}_{3 \times 3} & \textbf{I}_{3} & \textbf{0}_{3 \times 3} & \textbf{0}_{3 \times 3} & \textbf{0}_{3 \times 3}
    \end{bmatrix}.
    \end{equation}
    \item \textbf{Zero angular rate} \\
    Similarly to ZVU,  while in stationary conditions, the angular velocity vector in the reference frame is zero. The ZAR measurement residual is given by:
    \begin{equation} \label{eq:delta_z_ZAR}
    \delta \mathbf{z}_{_{\textbf{ZAR}}} = \boldsymbol{{\omega}}^b_{ib,_{{\text{INS}}}} - \textbf{0}_{3 \times 1} = \textbf{H}_{_{\textbf{ZAR}}} \delta \mathbf{x} + \boldsymbol{\nu}_{_{\textbf{ZAR}}} = \textbf{b}_g+ \boldsymbol{\nu}_{_{\textbf{ZAR}}}
    \end{equation}
    where $\boldsymbol{{\omega}}^b_{ib,_{{\text{INS}}}}$ is the measured INS angular rate vector, $\boldsymbol{\nu}_{_{\textbf{ZAR}}}$ is zero mean white Gaussian measurement noise,  and $\textbf{b}_g$ is the gyroscope bias vector expressed in the body frame. The corresponding measurement matrix is defined by:
    \begin{equation}\label{eq:HZAR}
    \textbf{H}_{_{\textbf{ZAR}}} = \begin{bmatrix}
    \textbf{0}_{3 \times 3} & \textbf{0}_{3 \times 3} & \textbf{0}_{3 \times 3} & \textbf{0}_{3 \times 3} & \textbf{I}_{3}
    \end{bmatrix}.
    \end{equation} 
\end{enumerate}
\subsection{Summary and Analytical Assessment}
\begin{figure}[htbp]
	\centering
    \includegraphics[width=0.5\textwidth]
    {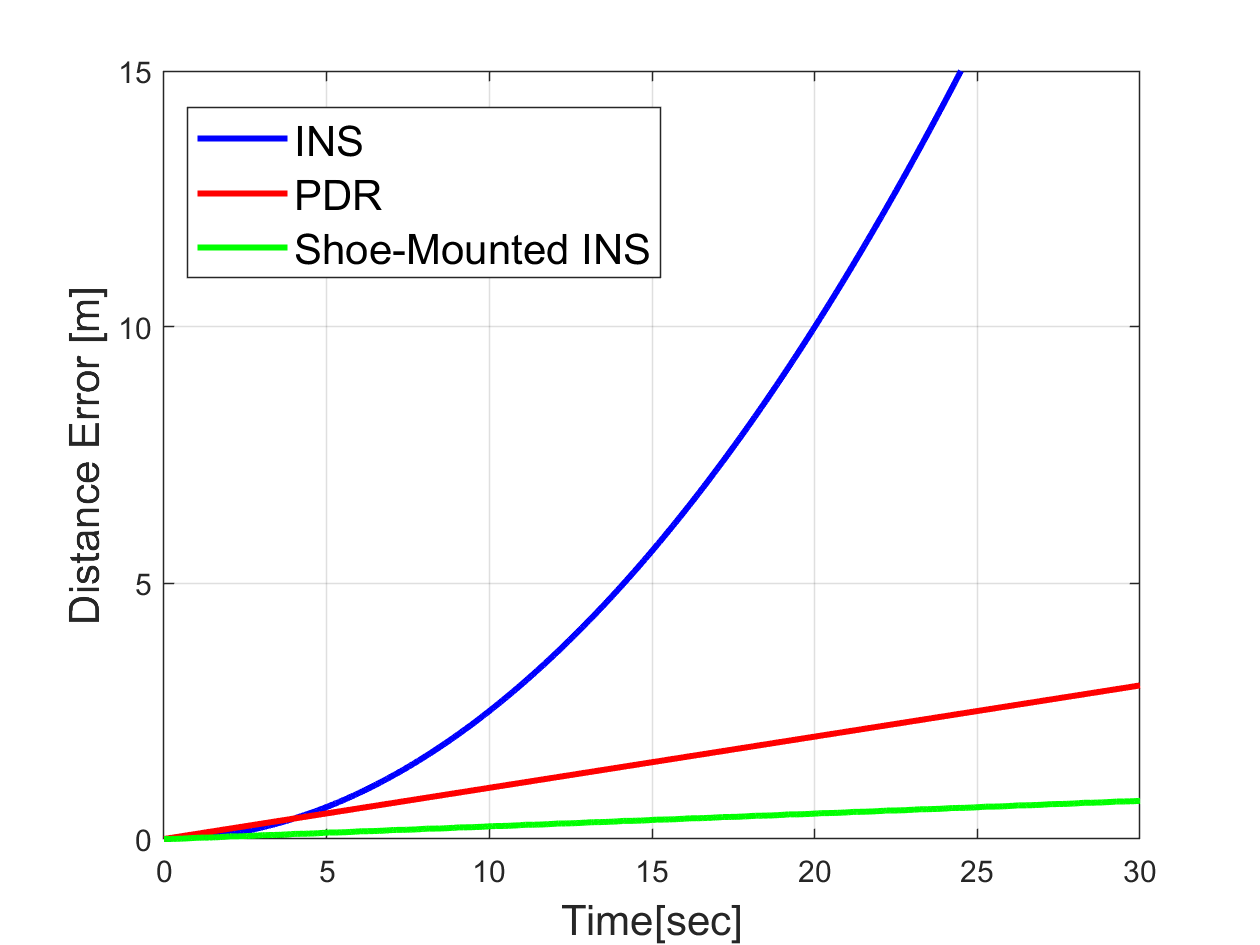}   
    \caption{Analytical assessment of the distance errors achieved by the INS, model-based PDR, and SM-INS.}\label{Fig:INDS_PDR_SMINS}
\end{figure}

\noindent
Algorithm~\ref{algo2} summarizes the SM-INS approach. The output of the algorithm is the updated error-state vector used to correct the INS solution and inertial sensor errors. \\
\begin{algorithm}[h!]
\caption{Shoe-mounted INS Algorithm}\label{algo2}
\textbf{Raw data} \KwIn{$\mathbf{f}^b,\boldsymbol{\omega}^b$}
\textbf{INS initial conditions} \KwIn{$\textbf{p}_k,\textbf{v}_k, \textbf{T}^{r}_{b,k}$}
\textbf{EKF} \KwIn{$\textbf{P}_k, \textbf{Q}, \textbf{F}, \textbf{G}, \delta\textbf{x}_k $}
\textbf{Information aiding} \KwIn{$ \textbf{R}, \textbf{H}, \delta\textbf{z}_k$}
\textbf{Zero velocity detectors} \KwIn{$\gamma$}
\textbf{INS:} Propagate INS equations of motion \eqref{eq:pade}-\eqref{eq:ins_posupdate} given initial conditions and inertial sensor readings\;
\textbf{EKF Prediction:} Given initial covariances, apply \eqref{eq:EKF1}-\eqref{eq:EKFend1} \;
\textbf{Zero velocity condition:} Employ three logical conditions \eqref{eq:C1}-\eqref{eq:C3}.
\If{stationary condition holds}{
   \textbf{EKF update}: Update phase \eqref{eq:EKF2}-\eqref{eq:EKFend2} using information aiding \eqref{eq:zzvu}-\eqref{eq:HZAR}};  
\KwOut{$\textbf{P}_{k+1},\delta\textbf{x}_{k+1}$}
\end{algorithm}
Next, we provide a simplified analytical assessment comparing the distance errors of the INS solution, model-based PDR, and SM-INS. For this, we assume that only an accelerometer bias of $b_a=5$mg is present in the system. For short time periods, the INS position error is approximated by \cite{titterton2004strapdown}:
\begin{equation}\label{eq:ins_pos_err}
    \delta p_{\text{INS}} = \frac{1}{2}b_a t^2
\end{equation}
where $t$ is the time.\\
To assess the distance error of model-based PDR, we follow \cite{etzion2023morpi}, which proposes the MoRPI framework, a horizontal PDR-like approach for mobile robots. It was shown that the position error using the Weinberg step-length approach \eqref{eq:steplength2} is the sum of all peak-to-peak distance errors  
\begin{equation}\label{eq:pdr_pos_err}
    \delta p_{\text{PDR}} = \sum^{N}_{j=1} \Delta k_w \Delta f_j
\end{equation}
where $N$ is the number of steps, $\Delta k_w$ is the error in the Weinberg gain, and $\Delta f_j$ is
\begin{equation}
    \Delta f_j = \left (f_{mag,max} - f_{mag,min} \right )^{1/4}
\end{equation}
For this example, we set $\Delta k_w=5\%$.\\
To produce the SM-INS distance error, the error was propagated using \eqref{eq:pdr_pos_err} during the filter prediction phase, and when zero velocity was detected, it was assumed that the filer manages to correct $90\%$ of the existing drift. In the examined scenario, stationary conditions were detected every one second. \\
Figure~\ref{Fig:INDS_PDR_SMINS} shows the distance error of the approaches for a duration of 30 seconds. Because of the inertial measurement errors and the dead-reckoning nature, the distance solution of all three approaches drift, although at different rates. The INS reaches a distance error of $22.5$m; the model-based PDR obtains a distance error of $3$m; and the SM-INS performs best, achieving a distance error of $0.75$m. This error represents an improvement of $75\%$ over the model-based PDR approach. In real world scenarios, this improvement may be higher, depending on several factors, including the walking duration.   
\section{Data-Driven Pedestrian Dead Reckoning}\label{sec:ddpdr}
\begin{figure}[htbp]
	\centering
    \includegraphics[width=0.5\textwidth]
    {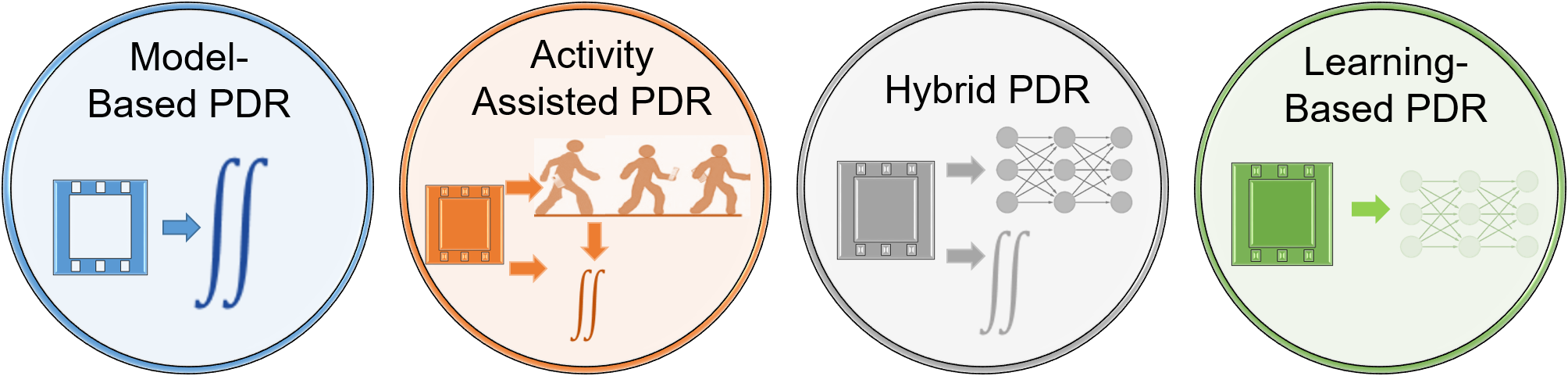}   
    \caption{Illustration of model-based PDR and three types of learning-based PDR approaches.}\label{Fig:types_of_pdr}
\end{figure}

Both machine learning and deep learning approaches learn from data, therefore they are referred to also as data-driven approaches. 
These approaches are used in a variety of navigation tasks~\cite{li2021inertial, klein2022data, chen2023deep, cohen2023inertial}, including PDR~\cite{wang2022recent}, as described below.
Data-driven PDR can be categorized into three types:
\begin{itemize}
    \item \textbf{Activity-assisted PDR}: Data-driven approaches are used to classify the user motion and inertial sensor location. This information is used within the model-based PDR algorithm.
    \item \textbf{Hybrid PDR approaches}: The heading angle or step-length are regressed by a data-driven approach while the other by using a model-based method.
     \item \textbf{Learning-based PDR}: Rather than being based on models, PDR positioning relies only on learning methods to perform the positioning task.
\end{itemize}
The above-mentioned categories as well as the model-based PDR are illustrated in Figure~\ref{Fig:types_of_pdr}. In the next section, a brief introduction to neural networks (NN) provides basic concepts for those who are unfamiliar with them.
\subsection{A Brief Introduction to Neural Networks}\label{sec:introN}
\begin{figure}[htbp]
	\centering
    \includegraphics[width=0.5\textwidth]
    {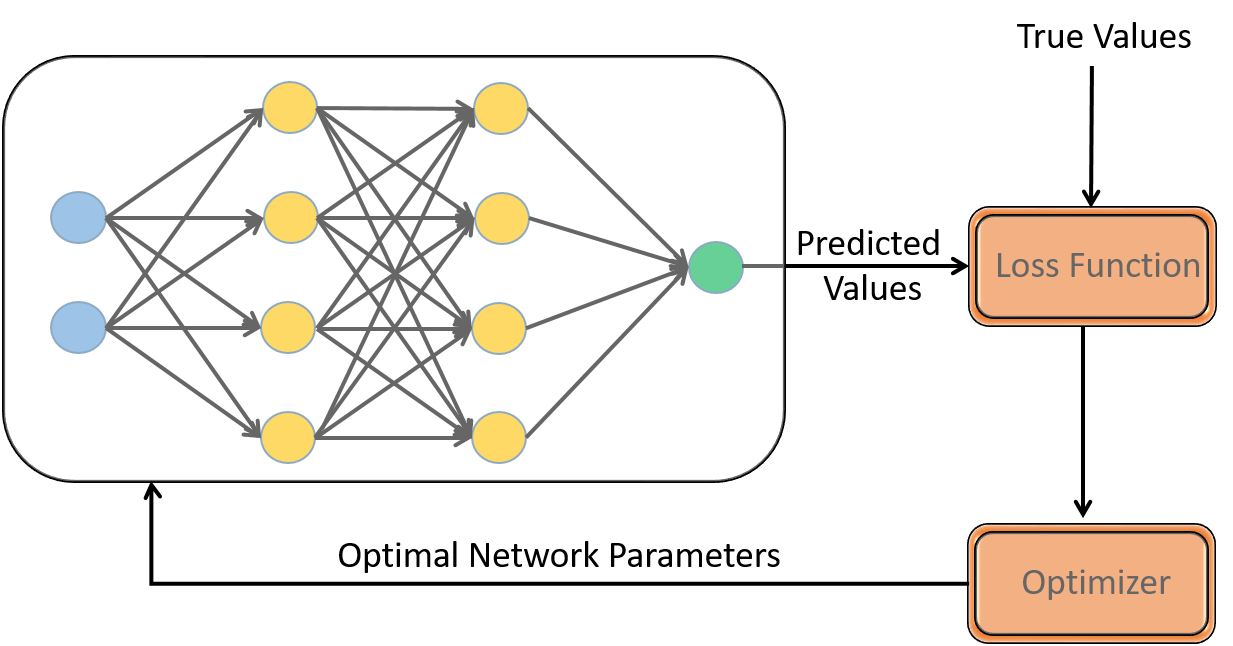}   
    \caption{A single cycle of the NN optimization process in its training phase.}\label{Fig:NN}
\end{figure}

\noindent
A typical NN consist of three types of layers: \textit{input layer}, holding the initial data for the entire network, \textit{output layer}, producing the result of the network, and \textit{hidden layer}, referring to all intermediate layers between the input and output layers. The output of the network is refereed to as the predicted value, which is compared to the true value. To this end, a loss function (cost function) is chosen to measure the difference between predicted and true values. Thus, the loss function is used as a criterion in the optimization process for obtaining undated network parameters. The optimization process is applied until the loss value is lower than a predefined threshold.  A cycle of this process is illustrated in Figure~\ref{Fig:NN}.
From a mathematical point of view, the j-th hidden layer $\textbf{h}_j$ is defined as
\begin{equation}
    \textbf{h}_j = {a} \left ( \sum^{m}_{i=0}x_i\times w_i \right )
\end{equation}
where $a$ is a nonlinear activation function, $x_i$ is the $i$, $i=0\ldots m$, input to the layer, and $w_i$ are the corresponding weights. The input layer $\textbf{l}_{\text{in}}\in\mathbb{R}^{n\times k}$ contains the input data of dimension $n$ to the network, where $k$ stands for the sequence length. For example, using the specific force vector for a sequence of 30 samples gives $n=3$ and $k=30$. The output layer, $\textbf{l}_{\text{out}}\in\mathbb{R}^{p}$, gives the output with dimension $p$. The complete network, \textbf{n}, is represented using a composition of functions, where each function is a layer in the network:    
\begin{equation}
    \textbf{n} = \textbf{l}_{\text{in}}\circ\textbf{h}_1\cdots\circ\textbf{h}_f\circ\textbf{l}_{\text{out}} 
\end{equation}
where $\textbf{h}_f$ is the last hidden layer in the network. Note that the network performed the following mapping:
\begin{equation}
    \textbf{n} : \mathbb{R}^{n\times k} \rightarrow \mathbb{R}^{p}
\end{equation}
A detailed explanation of NN can be found in \cite{goodfellow2016deep,zhang2023dive} and \cite{prince2023understanding}.
\subsection{Activity-Assisted PDR}
\begin{figure}[htbp]
	\centering
    \includegraphics[width=0.4\textwidth]
    {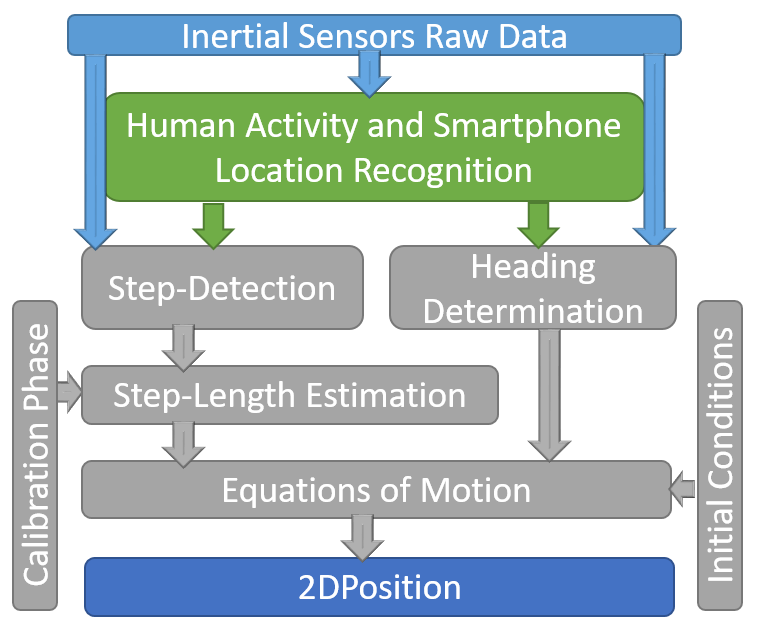}   
    \caption{Activity-assisted PDR stages. Machine learning algorithms are embedded in the model-based PDR for activity and location recognition.}\label{Fig:pdr_with_slr}
\end{figure}

\begin{figure}[htbp]
	\centering
    \includegraphics[width=0.4\textwidth]
    {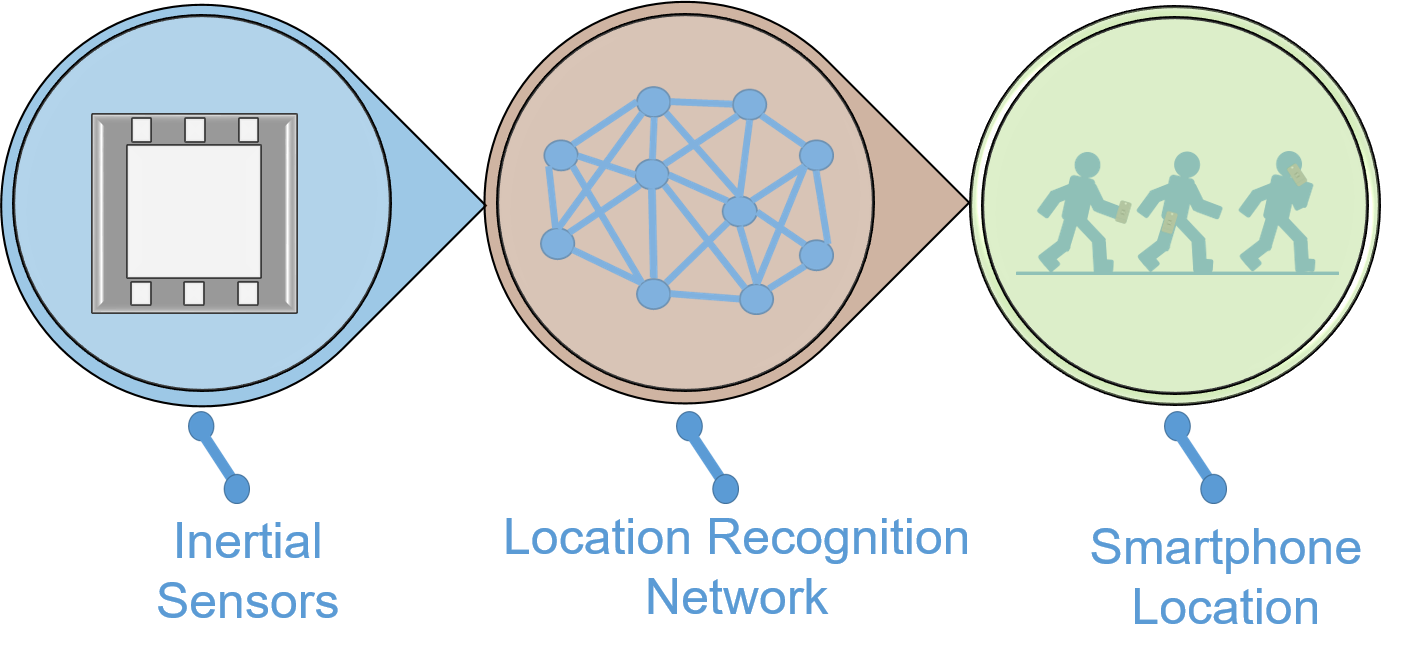}   
    \caption{Inertial sensors raw data is plugged into a smartphone recognition network to classify the smartphone location - texting, pocket, or talking.}\label{Fig:SLR_net}
\end{figure}

\noindent
One of the critical phases in model-based PDR is the step-length estimation. As addressed in Section~\ref{sec:steplegnth}, commonly, an empirical gain is required to estimate the step-length. This gain was found to be sensitive to the user activity mode (walking normally/slow/fast, standing, running)~\cite{elhoushi2016survey} and to the smartphone activity, which implies  the location of the inertial sensors \cite{klein2019smartphone}. For example, when using the smartphone sensors, the location could be texting, when writing a message, or pocket, when the smartphone is placed in the user's pocket. Each of such mode results in a different gain value for the step-length estimation phase. One solution to this problem is to take an average gain value of all expected walking mode and smartphone locations during the pedestrian trajectory. Yet, as shown in \cite{klein2018pedestrian}, even an average gain value results in a $10\%$ position error because of gain inaccuracy alone. To mitigate the influence of the human activity and sensor location, a dedicated machine learning algorithm is added to the model-based PDR stages, as illustrated in Figure~\ref{Fig:pdr_with_slr}, to create an activity-assisted PDR framework. 
Notice that, nevertheless, the burden of prior gain calibration is needed for all modes and sensor locations expected in the pedestrian trajectory. As soon as the classification learning algorithm identifies the current mode, the appropriate gain is selected for the PDR algorithm.
To train the learning algorithm, a data acquisition phase is conducted to collect inertial sensor data. Next, a preproccssing phase is applied, including handling missing data, normalization, noise reduction, and outlier rejection. If a machine-learning algorithm is used for classification, certain features are required. These features are extracted based on the specific force and angular velocity vector components. Feature types include statistical, time domain, cross-sensor, and frequency domain features. As the feature set is created and the required activity or location classes are defined, the training process is performed with a goal of finding the most efficient classifier for the given task. \\
The same data acquisition and prepossessing phases are used with deep learning approaches but the feature extraction part is not required because the network creates its own features from the raw inertial measurements. For classification problems, a commonly used loss function is the cross entropy loss, which evaluates the difference between the predicted and true class probability distributions.  \\
After the classifier has been designed, during the inference phase, raw inertial data (or features) are input to the classifier to output the human activity or the inertial sensor location. An example of this cycle is illustrated in Figure~\ref{Fig:SLR_net}, where raw inertial sensor data are plugged into a smartphone recognition network to classify the smartphone location texting, pocket, and talking. \\
See ~\cite{wang2022recentAR} to get a better understanding of recent advances in pedestrian-related activity recognition. Note that, as stated in~\cite{daniel2021smartphone}, supervised networks are trained on a set of defined user modes (smartphone locations or user activities), available during the training process. As a result, when the classifier encounters an unknown mode, it must identify it as one of the original modes that it was trained on. It is likely that such classification errors will reduce the accuracy of the PDR solution, therefore, appropriate algorithms should be applied to identify unknown modes.
\subsection{Hybrid PDR Approaches}
\noindent
Data-driven approaches to PDR were initially hybrid approaches. This means that they provide only one of the two quantities required for PDR. Those are the step-length and user heading. The second one is provided using a model-based PDR approach. Although recent approaches focus on a complete learning-based framework, as described in the next subsection, hybrid PDR still has utility because it requires a lighter computational load. PDR step-length estimation is of interest in healthcare and biomechanical applications. \\
\begin{enumerate}
    \item \textbf{Step-length learning}. Generally, data-driven approaches are aimed to regress the step-length estimation. They replace both the step detection and step-length estimation in a model-based procedure (Figure~\ref{Fig:pdr_model}), so that both the heading and the user position estimation are use model-based approaches, as illustrated in Figure~\ref{Fig:HybridStep}. In \cite{klein2020stepnet}, three StepNet architectures for the regression task are proposed. Two of them require the step detection phase and regress the step-length within the pedestrian step cycle (varying time intervals); the third, omits the step detection phase and regresses the change in distance at a predefined (constant) time interval. The last one demonstrated the best performance. Subsequently, the accuracy of step regression approaches was further improved by using magnetometer readings as well~\cite{bo2022mode}. 
    \begin{figure}[htbp]
	\centering
    \includegraphics[width=0.4\textwidth]
    {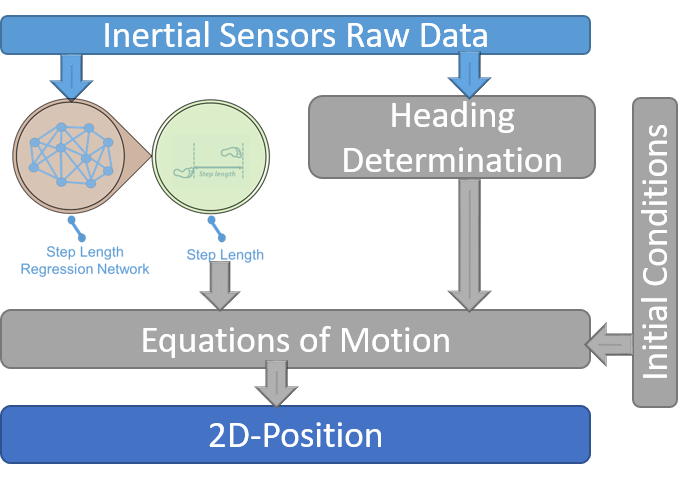}   
    \caption{Hybrid PDR with step-length data-driven methods.}\label{Fig:HybridStep}
    \end{figure}
    A variety of neural network architectures can be used for the regression task but these are beyond the scope of this paper.  Regardless of the architecture chosen, to solve the regression problem, the mean square error (MSE) loss is commonly applied:
    \begin{equation}\label{eq:mse_loss}
        \mathcal{L}_{\text{MSE}} = \frac{1}{N} \sum^{N}_{i=1} (\hat{x}_i - x_i)^2  
    \end{equation}
    where $N$ is the number of samples, $\hat{x}_i$ is the estimated value of sample $i$, and $x_i$ is the corresponding true value. In the training process, the network is optimized by minimizing the loss function. 
    \item \textbf{Heading learning}.
    Similarly, only the user heading can be regressed using data-driven approaches within the model-based PDR framework, as shown in Figure~\ref{Fig:HybridHeading}. As any other regression problem, the MSE loss \eqref{eq:mse_loss} can be used, but to better reflect the circular nature of heading, other loss functions may be considered. For example, in \cite{wang2019pedestrian}, the following heading loss was suggested:
    {\small
    \begin{equation}\label{eq:mse_loss_heading}
        \mathcal{L}_{\text{Heading}} = \frac{1}{N} \sum^{N}_{i=1} (\sin(\hat{\psi}_i) - \sin(\psi_i))^2+(\cos(\hat{\psi}_i) - \cos(\psi_i))^2  
    \end{equation} }
    where $\psi$ is the heading angle.
    \begin{figure}[htbp]
	\centering
    \includegraphics[width=0.4\textwidth]
    {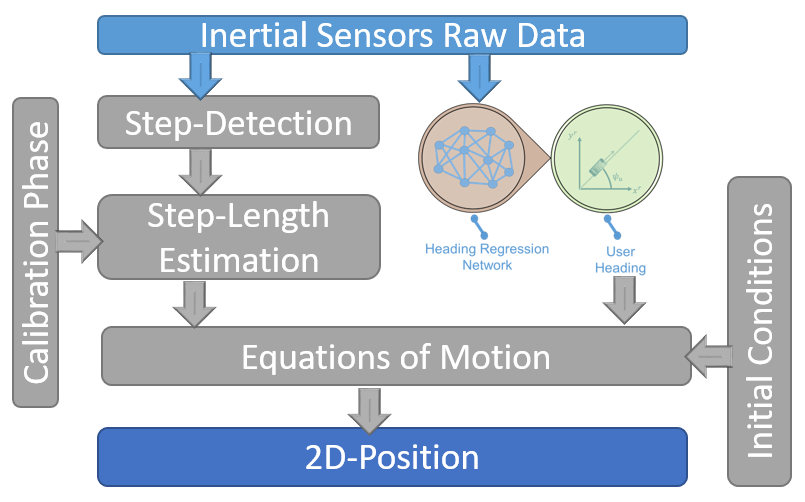}   
    \caption{Hybrid PDR with heading regression data-driven methods.}\label{Fig:HybridHeading}
    \end{figure}
\end{enumerate}
\subsection{Learning-Based PDR Frameworks}
\noindent
In learning-based PDR approaches, the entire model-based PDR algorithm is replaced by neural network architectures to estimate the pedestrian's position. In general, such approaches fall into two categories: 
\begin{enumerate}
    \item \textbf{Pedestrian position regression}. Following the same line of thought as in model-based PDR, such approaches regress the change in heading and distance to estimate the user position. IONet~\cite{chen2018ionet} examined LSTM and a bidirectional LSTM architecture, whereas \cite{asraf2021pdrnet} used residual networks (ResNets) for the regression only after the smartphone location was classified using a different network. \\
    As both the change in distance and heading are regressed simultaneously, the loss function accounts for both of them, as for example:
    \begin{equation}\label{eq:mse_loss2}
        \mathcal{L}_{\text{PDR}} = \frac{1}{N} \sum^{N}_{i=1} (\hat{d}_i - d_i)^2 +\lambda \frac{1}{N} \sum^{N}_{i=1} (\delta\hat{\psi}_i - \delta\psi_i)^2
    \end{equation}
    where $d$ is the distance, $\delta\psi$ is the change in heading, and $\lambda$ is a factor to balance the two losses. 
    \item \textbf{Pedestrian velocity regression}. This type of approach regresses the velocity of the user and integrates the data to determine the position of the user based on that velocity. The first work in this field, named RIDI~\cite{yan2018ridi}, regressed the user velocity and used it to correct the user acceleration. Next, double integration was applied to the corrected acceleration to estimate the pedestrian position.  By building upon RIDI, RoNIN~\cite{herath2020ronin} offers a heading-agnostic coordinate frame representing the input and output of the network. They examined three different network architectures based on ResNet, LSTM, and temporal convolutional layers. In contrast to other approaches, RoNIN uses the device orientation together with the inertial readings to provide input to the network, as opposed to using only the inertial readings.     
\end{enumerate} 
\section{Summary}\label{sec:conc}
Inertial navigation for pedestrians is an emerging discipline that has wide applications in many fields, including healthcare, security, and indoor location-based services. 
In this work, we presented model and learning approaches to inertial pedestrian navigation. We provided detailed algorithms for shoe-mounted inertial sensors and classical PDR with unconstrained inertial sensors. These algorithms include the INS equations of motion and the EKF framework aided by information in stationary conditions. For PDR, we described methods for solving each phase of the algorithm for step detection, step-length estimation, and heading determination. Following these model-based approaches, we addressed three categories of data-driven PDR: activity-assisted and hybrid approaches, and learning-based frameworks.
\bibliographystyle{IEEEtran}
\bibliography{ref}

\end{document}